\pdfoutput=1

\documentclass[11pt]{article}

\usepackage[final]{acl}

\usepackage{times}
\usepackage{latexsym}
\usepackage{booktabs}
\usepackage{makecell}
\usepackage{array}
\usepackage{enumerate}
\usepackage{enumitem}

\usepackage{xspace}

\usepackage[T1]{fontenc}

\usepackage[utf8]{inputenc}

\usepackage{microtype}

\usepackage{inconsolata}

\usepackage{graphicx}
\usepackage{booktabs}
\usepackage{multirow}
\usepackage{multicol}
\usepackage{cleveref}
\usepackage{tcolorbox}
\usepackage{xcolor}
\usepackage{listings}
\usepackage[normalem]{ulem}
\usepackage{ragged2e}
\newcommand{\gw}[1]{\textcolor{blue}{\small [GW: #1]}}
\newcommand{\emily}[1]{\textcolor{orange}{\small [EA: #1]}}
\newcommand{\xt}[1]{\textcolor{purple}{\small [XT: #1]}}
\definecolor{mypurple1}{RGB}{143, 94, 255}
\newcommand{\cy}[1]{\textcolor{mypurple1}{\small [Cy: #1]}}
\newcommand{\rs}[1]{\textcolor{magenta}{\small [RS: #1]}}

\renewcommand{\gw}[1]{}
\renewcommand{\emily}[1]{}
\renewcommand{\xt}[1]{}
\renewcommand{\cy}[1]{}
\renewcommand{\rs}[1]{}

\title{Analyzing LLM Instruction Optimization for Tabular Fact Verification}

\author{
\normalsize
Xiaotang Du$^1$\quad
Giwon Hong$^1$\quad
Wai-Chung Kwan$^1$\quad
Rohit Saxena$^1$\quad
Ivan Titov$^1$\\
\normalsize
\textbf{Pasquale Minervini}$^{1,2}$\quad
\textbf{Emily Allaway}$^1$
\\
\normalsize
$^1$University of Edinburgh, United Kingdom \qquad
$^2$Miniml.AI, United Kingdom\\
\normalsize
\texttt{\{xiaotang.du, ghong, p.minervini, emily.allaway\}@ed.ac.uk}\\
}

\begin{document}
\maketitle

\begin{abstract}
Instruction optimization provides a lightweight, model-agnostic approach to enhancing the reasoning performance of large language models (LLMs).
This paper presents the first systematic comparison of instruction optimization, based on the DSPy optimization framework, for tabular fact verification.
%
We evaluate four out-of-the-box prompting techniques 
that cover both text-only prompting and code use: direct prediction, Chain-of-Thought (CoT), ReAct with SQL tools, and CodeAct with Python execution.
We study three optimizers from the DSPy framework---COPRO, MiPROv2, and SIMBA---
across four benchmarks
and three model families.
%
%
We find that instruction optimization consistently improves verification accuracy, with MiPROv2 yielding the most stable gains for CoT, and SIMBA providing the largest benefits for ReAct agents, particularly at larger model scales.
Behavioral analyses reveal that SIMBA encourages more direct reasoning paths by applying heuristics, thereby improving numerical comparison abilities in CoT reasoning and helping avoid unnecessary tool calls in ReAct agents.
Across different prompting techniques, CoT remains effective for tabular fact checking, especially with smaller models.
Although ReAct agents built with larger models can achieve competitive performance, they require careful instruction optimization.
\end{abstract}

\section{Introduction}
Verifying natural language claims against structured data is a central capability for trustworthy NLP systems deployed in science, public health, and information quality assurance.
While numerous methods have been proposed for tabular fact verification~\cite[\textit{inter alia}]{yang-zhu-2021-exploring-decomposition,ou-liu-2022-learning,lu-etal-2025-tart,zhang2024reactable}, the resulting systems are often specialized to a particular dataset or fail to outperform simpler prompting approaches.


In this work, we conduct a comparative study of out-of-the-box prompting techniques, paired with instruction optimization, for tabular fact verification. 
Instruction optimization is a technique that allows for improvements to LLM performance without gradient updates. 
Since LLMs are known to be sensitive to prompt formulation~\cite{webson-pavlick-2022-prompt,leidinger-etal-2023-language}, we analyze the impacts of instruction optimization on
practical and generalizable prompting techniques, such as Chain-of-Thought~\cite{wei2022chain}.
%


Recent frameworks for instruction optimization (e.g., DSPy;~\citealp{khattab2023dspy}) treat multi-step LLM pipelines as programs whose textual parameters can be automatically tuned by search or meta-reasoning, yielding large gains on diverse tasks. Despite this progress, a systematic understanding of how instruction optimization affects tabular fact verification is lacking. 
The following impacts are particularly underexplored:
(1) prompting techniques that differ in their intermediate computation (e.g., direct prediction, CoT, and program-aided reasoning via SQL and Python), (2) optimizer families, and (3) model scale and families. Tool-augmented agents (e.g., ReAct;~\citealp{yao2023react}) promise stronger grounding by interleaving thoughts with executable actions, but their end-to-end effectiveness depends critically on the learned tool interface and execution reliability—factors that instruction optimization may help or hinder.

We present the first comparative study of instruction optimization for tabular fact verification using the DSPy optimization framework.
Our study focuses on three optimizers within DSPy: COPRO, MiPROv2, and SIMBA\footnote{We restrict MiPROv2 and SIMBA to instruction-only tuning to isolate the effect of instructions from few-shot example selection.}.
We analyze these optimizers across
four benchmarks (TabFact, PubHealthTab, SciTab and MMSci), 
four prompting techniques
(Direct prediction, CoT, ReAct, and CodeAct), 
and three base LLMs (Qwen3, Gemma3, GPT-4o). 
%
We conduct a comprehensive analysis, using both task metrics (accuracy, macro-F1) and behavioral shifts (e.g., frequency and success of SQL/Python calls, error taxonomies), to address the following research questions:
\begin{itemize}[itemsep=0em,leftmargin=10pt]
    \item What is the impact of optimized instructions on CoT reasoning?
    \item How does the optimized instructions affect the tool calling behavior of ReAct agent?
    \item Does program-aided reasoning show consistent advantages over CoT in tabular fact checking? 
\end{itemize}

Our experiment results show that the influence of instruction optimization varies depending on the prompting technique and model size. 
Compared with smaller models, larger models benefit more from the optimized instructions and show greater improvement across four test sets.
Among different optimizers, MiPROv2 brings consistent gains for CoT reasoning, while SIMBA proves effective for improving tool use behavior of ReAct. 

A closer inspection of the optimized instructions reveal that MiPROv2 and SIMBA optimizer encourage the model to choose a direct reasoning path, avoiding unnecessary tool calls for simple claims.
With SIMBA optimizer, the tool use frequency is reduced, with more successful tool execution.
In addition, we find SIMBA optimizer includes more heuristics about numerical comparison in the refined instructions, which helps improve the overall accuracy in CoT.
Our code is available at \url{https://github.com/xiaodu0123/tabfact-prompt-optimization}.
%






\section{Related Work}
\label{sec:related}
\paragraph{Table-based Fact Checking}
Verifying claims against structured evidence requires compositional reasoning over diverse table schema. TabFact~\cite{ChenWCZWLZW20} established the first large-scale benchmark for binary fact verification on Wikipedia tables.
Later datasets incorporated more nuanced labeling schema (e.g., three labels instead of only two) and more complex claims requiring multi-hop reasoning~\cite{wang-etal-2021-semeval}. Among these, several domain-specific datasets have been created: PubHealthTab~\cite{akhtar-etal-2022-pubhealthtab}, which targets claims about public health, SciTab~\cite{lu2023scitab}, which includes claims from computer science publications, and SciAtomicBench~\cite{zhang2025atomic}, which covers computer science along with other domains such as finance. 
While fact verification datasets typically present tabular data in textual form, multi-modal datasets have also been created~\cite{yang2025does}.
Additionally, some fact-verification datasets mix tabular evidence with text~\cite{aly-etal-2021-fact,schlichtkrull2023averitec,zhao2024findver} and figures~\cite{wang-etal-2025-sciver,chan-etal-2024-overview}.

Early methods for tabular fact verification used symbolic or programmatic reasoning~\cite{ChenWCZWLZW20,zhong-etal-2020-logicalfactchecker,shi-etal-2020-learn,zhang-etal-2020-table,yang-etal-2020-program,yang-zhu-2021-exploring-decomposition,ou-liu-2022-learning}. While some recent work has also made use of neuro-symbolic systems~\cite{glenn-etal-2024-blendsql,aly-vlachos-2024-tabver,Cheng2022BindingLM}, there has been an increasing focus on adapting and making use of LLMs. To this end, prior works have developed both pre-training~\cite{eisenschlos-etal-2020-understanding,dong-smith-2021-structural,zhang-etal-2024-tablellama} and fine-tuning~\cite{wu-feng-2024-protrix,jiang2025tablemind} approaches for table-based fact verification, as well as more general table-based reasoning tasks~\cite{herzig-etal-2020-tapas,Liu2021TAPEXTP}. Additionally, several works propose prompting techniques for improving model reasoning over tables~\cite{wang2024chain,zhang2025atomic,abhyankar-etal-2025-h,zhang2024reactable}. Recently, work has also begun to investigate agentic systems and tool-use for table-based fact verification~\cite{lu-etal-2025-tart,zhou-etal-2025-efficient}. However, despite these advances, many systems are computationally intensive or specialized to a particular dataset. In contrast, our work explores computationally light instruction optimization techniques applied to general prompting strategies.

Most closely related to our work are two recent analyses into the challenges of various table understanding tasks, including fact verification. \citet{bhandari2024exploring} examine how instruction tuning, in-context examples, and model size impact performance on tabular reasoning tasks, while \citet{wu2025tabular} survey approaches to table understanding tasks more broadly. In contrast to these analyses, our work compares instruction optimization techniques applied to simple prompting strategies (standard baselines such as CoT as well as simple programmatic reasoning models such as ReAct). Additionally, while \citet{bhandari2024exploring} cover multiple table understanding tasks, our work focuses only on table-based fact verification, opting instead to cover a wider range of datasets tabular fact verification.

\section{Method}
\subsection{Prompting Techniques}
\paragraph{Chain-of-Thought} Chain-of-thought reasoning (CoT) \cite{wei2022chain} encourages LLMs to generate intermediate reasoning steps before producing the final answer. 
With CoT, LLMs can decompose a complex query into sub-problems and progressively build the solution in the reasoning traces.
%
\paragraph{ReAct} ReAct \cite{yao2023react} serves as a foundational framework for tool-based agents by interleaving reasoning with task-specific actions.
ReAct enables LLMs to interact with external tools, allowing them to collect additional evidence and ground their reasoning in the tool execution output.
In our experiments, we evaluate a ReAct agent with access to a standard SQL tool that can execute SQL queries on the table data to retrieve relevant information and perform math operations.

\paragraph{CodeAct}
CodeAct \cite{wang2024executable} leverages executable Python code as the primary action modality for tool-based agents.
Unlike existing paradigms that rely on tool calls in text or JSON formats, CodeAct enables multi-step operations and flexible tool chaining through code execution, allowing the agent to perform sophisticated actions by integrating with Python's control flow and existing libraries.
In our experiments, the CodeAgent has no access to pre-defined tools.
It generates free-form python codes to process the table data and perform math operations step by step.

\subsection{Instruction Optimization}
In our analysis, we focus on three LLM-based instruction optimization approaches in the DSPy~\cite{khattab2023dspy} framework:
COPRO, MiPROv2 \cite{opsahl-ong-etal-2024-optimizing} and SIMBA.
%

\paragraph{DSPy Framework}
DSPy is a framework for algorithmically optimizing model prompts and weights, treating LLM pipelines as programmes that can be automatically compiled and optimized.

\paragraph{COPRO} Cooperative Prompt Optimization (COPRO) systematically explores various candidate instructions in a beam search-like manner and evaluates their performance on the train set. 
The optimizer iteratively refines the prompt instruction by proposing multiple new candidate instructions based on the N best prompts among previous attempts and their corresponding evaluation scores.

\paragraph{MiPROv2} Multi-Stage Instruction Prompt Optimization (MiPROv2) is an advanced framework that can refine both the instruction and few-shot demonstrations through a three-stage pipeline.
First, the optimizer bootstraps multiple candidate sets of few-shot demonstrations from the training data.
Then, it generates diverse prompt instructions and demonstrations based on previously evaluated candidates, the properties of the downstream task, and randomly sampled prompting strategies.
Finally, MiPROv2 employs Bayesian optimization method to efficiently search the best combination of candidate instruction and demonstration. 

Compared with COPRO, MiPROv2 provides a richer context for the generation of new candidate instructions and performs more efficient evaluation on mini-batches of training data.

\paragraph{SIMBA} Stochastic Introspective Mini-Batch Ascent (SIMBA) is an introspective prompt optimization algorithm that leverages the language model's capacity for self-reflection to iteratively improve instruction quality. 
The optimizer identifies challenging training instances where model outputs exhibit high variability, then applies two complementary strategies to refine prompts. 
One strategy performs contrastive analysis, where the model compares successful and unsuccessful execution traces to generate explicit improvement rules that are appended to the original instruction. 
Another strategy incorporates successful execution trajectories as few-shot demonstrations. 

\section{Experiments}

\subsection{Datasets}
We evaluate the performance of various LLMs on four tabular fact checking datasets: TabFact~\cite{ChenWCZWLZW20}, PubHealthTab~\cite{akhtar-etal-2022-pubhealthtab}, SciTab~\cite{lu2023scitab} and MMSci~\citep{yang2025does}. 
These datasets cover diverse domains and table types, ranging from general knowledge to specialized data, thereby enabling a more comprehensive evaluation of the generalization ability of different approaches.
In SciTab, PubHealthTab, and MMSci,
there are three labels:
\textit{supports}, \textit{refutes} and \textit{not enough info}; TabFact 
is 
a binary classification task with only \textit{supports} and \textit{refutes} labels.


\paragraph{SciTab} SciTab~\cite{lu2023scitab} is a benchmark designed for scientific claim verification, leveraging real-world table evidence from scientific publications in machine learning and natural language processing domains.
The dataset presents unique challenges in claim ambiguity, compositional reasoning and numerical analysis of scientific data.

\paragraph{PubHealthTab} PubHealthTab~\cite{akhtar-etal-2022-pubhealthtab}  is a table-based fact checking dataset focusing on public health claims.
The evidence tables are extracted from multiple web sources, which exhibit noisy and complex table structure with varying content quality.

\paragraph{TabFact} TabFact~\cite{ChenWCZWLZW20} is a large-scale table-based fact verification dataset that consists of human-annotated claims with Wikipedia tables as evidence. 
TabFact provides two test sets that differ in the claim complexity, and we use the complex test set for evaluation.
\paragraph{MMSci}
MMSci~\citep{yang2025does} is a benchmark for multimodal scientific reasoning across three table-based tasks. 
We use the table fact verification test set, converting table images to textual format,
%
to evaluate generalization 
to unseen data.


\subsection{Optimization}
For each considered LLM, we evaluate the performance of different prompting techniques, including direct prompting, CoT, ReAct and CodeAct to study the impact of instruction optimization on both language-based reasoning and program-aided reasoning.
%
We use the same instructions in the system prompt before optimization for different experiments, i.e. \texttt{verify the given claim against the provided table data.} 
%
All the experiments are conducted in zero-shot setting.

\begin{table}[!t]
\centering
\small
\begin{tabular}{lrrr}
\toprule
\textbf{Dataset} & \textbf{Train} & \textbf{Dev} & \textbf{Test} \\
\midrule
TabFact & 92,585 & 12,851 & 8,609 \\
SciTab & 210 & 429 & 429 \\
PubHealthTab & 1440 & 177 & 180 \\
MMSci & - & - & 1,038 \\
\bottomrule
\end{tabular}
\caption{Statistics of the 
fact checking datasets.}
\label{tab:dataset_stats}
\end{table}

\subsection{Evaluation Setup}
\paragraph{Models and Baselines} We conduct our experiments using Qwen3 \cite{qwen3}, Gemma3 \cite{gemmateam2025gemma3technicalreport} and GPT-4o models, which allows us to systematically investigate the impact of instruction optimization on reasoning and tool-calling behavior across different model families and sizes.
The same model is used for proposing candidate instructions and evaluating instruction quality during optimization.
To examine the effectiveness of optimized instructions, we compare the model performance in GPT-4o experiments with ReActable \cite{zhang2024reactable}, a ReAct framework that uses GPT-4o with human-written instructions and 
SQL and Python as tools.

\paragraph{Data processing}
Each fact checking dataset is processed into
a unified data format.
%
We then split three of the datasets (TabFact, SciTab, and PubHealthTab) into train, development and test sets; our fourth dataset, \textit{MMSci, is used only for evaluation}.
We create a hybrid training set for instruction optimization by randomly sampling 100 instances from the training splits of the three datasets.
We sample 40 PubHealthTab instances, 40 SciTab instances and 20 TabFact instances to ensure the label distribution of the hybrid dataset is balanced.
Statistics of the processed datasets are in Table \ref{tab:dataset_stats}.
%

%


\paragraph{Evaluation metrics}
We optimize the instructions 
using
the hybrid train data, and evaluate the performance on the development and test sets of all four datasets with
%
accuracy and macro-F1.
During instruction optimization, only accuracy is used to measure the quality of different candidate prompts.

\begin{table*}[th!]
\centering
\resizebox{\textwidth}{!}{
\begin{tabular}{llcccccccccccccccc}
\toprule
& & \multicolumn{8}{c}{\textbf{Qwen3-8B}} & \multicolumn{8}{c}{\textbf{Qwen3-32B}} \\
\cmidrule(lr){3-10} \cmidrule(lr){11-18}
\textbf{Module} & \textbf{Optimizer} & \multicolumn{2}{c}{\textbf{PubHealth}} & \multicolumn{2}{c}{\textbf{SciTab}} & \multicolumn{2}{c}{\textbf{TabFact}} & \multicolumn{2}{c}{\textbf{MMSci}} & \multicolumn{2}{c}{\textbf{PubHealth}} & \multicolumn{2}{c}{\textbf{SciTab}} & \multicolumn{2}{c}{\textbf{TabFact}} & \multicolumn{2}{c}{\textbf{MMSci}} \\
\cmidrule(lr){3-4} \cmidrule(lr){5-6} \cmidrule(lr){7-8} \cmidrule(lr){9-10} \cmidrule(lr){11-12} \cmidrule(lr){13-14} \cmidrule(lr){15-16} \cmidrule(lr){17-18}
& & Acc & F1 & Acc & F1 & Acc & F1 & Acc & F1 & Acc & F1 & Acc & F1 & Acc & F1 & Acc & F1 \\
\midrule
\multirow{4}{*}{Direct} 
& Baseline & \textbf{73.3} & \textbf{73.4} & \textbf{58.7} & 56.5 & 58.0 & 52.8 & 57.2 & \textbf{41.5} & 84.4 & 82.3 & 52.9 & 49.6 & 64.1 & 62.8 & 68.2 & 46.3 \\
& +COPRO & \textbf{73.3} & \textbf{73.4} & \textbf{58.7} & \textbf{56.6} & 58.1 & 52.8 & 57.0 & 41.2 & 84.4 & 82.7 & \textbf{53.4} & \textbf{51.1} & 65.3 & 63.9 & 69.5 & 47.7 \\
& +MiPROv2 & 72.8 & 72.9 & 56.4 & 54.3 & \textbf{58.5} & \textbf{53.4} & 56.6 & 41.3 & 84.4 & 82.3 & 53.1 & 50.0 & 64.1 & 62.8 & 68.2 & 46.3 \\
& +SIMBA & \textbf{73.3} & \textbf{73.4} & \textbf{58.7} & 56.5 & 58.1 & 52.8 & \textbf{57.3} & 41.4 & \textbf{85.6} & \textbf{84.3} & 52.7 & 50.0 & \textbf{67.6} & \textbf{68.7} & \textbf{70.6} & \textbf{49.4} \\
\midrule
\multirow{4}{*}{CoT} 
& Baseline & 83.9 & 82.3 & 64.3 & 64.4 & 77.6 & 80.5 & 81.9 & 58.0 & 88.3 & 87.6 & 66.4 & 66.4 & 84.5 & 86.6 & 86.5 & 61.6 \\
& +COPRO & 83.9 & 82.8 & \textbf{66.2} & \textbf{66.2} & 76.8 & 79.9 & 79.4 & 56.3 & 87.2 & 86.1 & 67.4 & 67.3 & 85.5 & 87.6 & 86.7 & 61.5 \\
& +MiPROv2 & \textbf{86.1} & \textbf{85.7} & 66.0 & 66.0 & \textbf{80.3} & \textbf{83.1} & \textbf{82.5} & \textbf{59.4} & 87.2 & 86.5 & \textbf{68.8} & \textbf{68.6} & \textbf{86.9} & \textbf{88.5} & \textbf{87.7} & \textbf{65.4} \\
& +SIMBA & 82.2 & 81.4 & 62.5 & 62.2 & 77.6 & 80.6 & 81.6 & 58.7 & \textbf{90.0} & \textbf{89.6} & \textbf{68.8} & \textbf{68.6} & 85.2 & 87.1 & 87.0 & 64.2 \\
\midrule
\multirow{4}{*}{ReAct} 
& Baseline & \textbf{86.7} & \textbf{86.6} & 61.3 & 61.2 & \textbf{83.8} & \textbf{85.3} & 82.5 & 58.5 & 87.8 & 87.4 & 61.5 & 60.1 & \textbf{86.4} & \textbf{87.0} & \textbf{87.5} & 62.6 \\
& +COPRO & 84.4 & 83.9 & \textbf{62.2} & \textbf{62.1} & 80.5 & 81.5 & \textbf{83.6} & \textbf{61.5} & 86.1 & 84.7 & 62.0 & 61.5 & 81.5 & 84.1 & 85.0 & 61.0 \\
& +MiPROv2 & 81.7 & 81.1 & 61.8 & 61.8 & 75.5 & 80.6 & 82.2 & 60.0 & 87.8 & 87.2 & 61.5 & 60.9 & 84.2 & 85.2 & 86.2 & 63.0 \\
& +SIMBA & 86.1 & 85.2 & 58.3 & 58.3 & 82.9 & 84.7 & 80.8 & 57.3 & \textbf{90.6} & \textbf{90.0} & \textbf{66.2} & \textbf{65.9} & 86.1 & \textbf{87.0} & 85.9 & \textbf{65.0} \\
\midrule
\multirow{4}{*}{CodeAct} 
& Baseline & \textbf{86.1} & \textbf{86.0} & 57.1 & 57.1 & 82.0 & 83.5 & 81.2 & 59.2 & 85.6 & 84.9 & 58.0 & 57.5 & 85.9 & 87.1 & 87.5 & \textbf{66.1} \\
& +COPRO & 82.8 & 82.2 & \textbf{59.7} & 59.1 & 80.0 & 82.2 & 83.3 & \textbf{60.7} & \textbf{87.2} & \textbf{86.6} & 62.2 & 61.8 & \textbf{86.7} & \textbf{87.9} & \textbf{88.1} & 63.3 \\
& +MiPROv2 & \textbf{86.1} & 85.7 & 56.9 & 56.6 & 80.5 & 82.0 & 82.1 & 59.0 & 83.9 & 83.5 & 59.0 & 58.2 & 86.4 & 87.6 & 86.5 & 62.7 \\
& +SIMBA & 85.0 & 84.9 & \textbf{59.7} & \textbf{59.5} & \textbf{84.8} & \textbf{85.5} & \textbf{84.3} & 59.8 & 85.6 & 85.2 & \textbf{69.2} & \textbf{69.3} & 85.4 & 87.0 & 86.5 & 62.6 \\
\bottomrule
\end{tabular}
}
\caption{Results of Qwen3-8B and Qwen3-32B on test sets. \textbf{Bold} is best performance per method and dataset. \xt{TODO: evaluate ReActable with Qwen3 models}}
\label{tab:qwen3_combined_results}
\end{table*}

\section{Results}
\xt{TODO: double check the claims based on new results; it might be hard to conclude what works best for CodeAct}
We report the test performance of different prompting techniques with Qwen3 models on four fact checking datasets in Table \ref{tab:qwen3_combined_results}. 
For direct prompting and CoT, larger models generally achieve higher accuracy and F1 than their smaller counterparts across most test sets.
For program-aided reasoning paradigms (ReAct, CodeAct), increasing model size does not yield significant performance gains. 
Although larger models have similar baseline performance to smaller versions, they benefit substantially more from instruction optimization and show greater improvement with refined instructions.

The effectiveness of instruction optimization for tabular reasoning is highly dependent on both model scale and the prompting technique.
For optimizing CoT reasoning, MiPROv2 brings the most consistent gains, achieving the highest accuracy and F1 on PubHealthTab, TabFact and MMSci for Qwen3-8B, and showing competitive results across three datasets with Qwen3-32B.
For program-based reasoning, SIMBA provides the strongest performance gain on SciTab, particularly for improving ReAct with the Qwen3-32B model. %
COPRO also offers moderate benefits for Qwen3-32B model but less consistently than SIMBA. 
This suggests that larger models are better at identifying patterns of successful trajectories through self-reflection and comparative analysis, leading to more effective rules for optimizing tool use in diverse scenarios.
\xt{TODO: provide qualitative analysis on SIMBA instructions as evidence?}

According to Table \ref{tab:gemma3_combined_results}, the general trend observed with the Gemma3 model family is slightly different from Qwen3. The larger Gemma3 model shows consistently higher performance for both CoT reasoning \textit{and} program-aided reasoning. 
Unlike Qwen3, where the optimizers fail to enhance the performance for ReAct with a smaller model, Gemma3 models respond more positively to instruction optimization across different prompting techniques and show greater improvement with refined instructions at both sizes.

Similar to Qwen3 experiments, MiPROv2 still delivers significant improvements when optimizing CoT. SIMBA performs exceptionally well for improving ReAct and CodeAct, particularly for the larger 27B model.
COPRO remains effective for smaller model (12B) but provides smaller incremental gains relative to MiPROv2 and SIMBA.
Overall, the Gemma3 model family underperforms Qwen3, even after applying instruction optimization.
For both Gemma3 and Qwen3 models, CoT reasoning consistently achieves competitive performance after instruction optimization compared with program-aided reasoning methods on tabular fact checking.

\Cref{tab:gpt4_combined_results_tabfact_mini} summarizes the test performance of GPT-4o models.
%
Due to budget considerations, GPT-4o models and ReActable are evaluated on a smaller TabFact test set (TabFact-mini) with 400 random instances. 
GPT-4o models demonstrate much stronger baseline performance, and consequently benefit less from instruction optimization than Qwen3 and Gemma3 models. 
For GPT-4o-mini, MiPROv2 is more effective for improving CoT reasoning, while SIMBA yields greater improvements across the test sets for optimizing ReAct. 
However, no single optimizer provides consistent performance gains for optimizing CodeAct.
For the GPT-4o model, SIMBA performs consistently well and brings improvement to both CoT and ReAct, whereas MiPROv2 is shown to be effective for enhancing CodeAct performance.
ReAct with GPT-4o shows slightly worse performance on SciTab and TabFact-mini compared with the ReActable baseline, but it can 
consistently outperform ReActable across all test sets after SIMBA optimization, which demonstrates the superiority of DSPy-based instruction optimization over manually designed prompts.

\xt{!Check the new section on MMSci =>}
According to the test performance on MMSci, we observe that for Qwen3-32B and Gemma3-27B model, the optimized instructions with superior performance on PubHealthTab, SciTab and TabFact often generalize well to MMSci.
Specifically, instructions optimized by SIMBA consistently achieves the highest F1 scores on MMSci in both direct prompting and ReAct settings, while CoT instructions learned by MiPROv2 continues to deliver the strongest improvements on MMSci.
However, this trend is not observed in GPT-4o models, for which the performance on the other three fact checking datasets is not predictive of test performance on MMSci. 
Although SIMBA shows strong performance on SciTab and TabFact-mini across direct prompting, CoT and ReAct settings, these performance gains do not consistently transfer to MMSci test data.
This may indicate instructions proposed by GPT-4o during SIMBA optimization generalize less effectively on unseen data.

To further examine the effectiveness and generalizability of instruction optimization, we conduct ablation studies using varying random seeds, initial instructions of diverse quality and different training data. 
We also extend the evaluation of ReAct agents by considering single or multiple most commonly used tools introduced in TART framework \cite{lu-etal-2025-tart}. \xt{Do I need to summarize some main findings from ablation study here?}
More detail of our experiments can be found in \Cref{sec:ablation_study}.
%



\begin{table*}[ht!]
\centering
\resizebox{\textwidth}{!}{
\begin{tabular}{llcccccccccccccccc}
\toprule
& & \multicolumn{8}{c}{\textbf{Gemma3-12B}} & \multicolumn{8}{c}{\textbf{Gemma3-27B}} \\
\cmidrule(lr){3-10} \cmidrule(lr){11-18}
\textbf{Module} & \textbf{Optimizer} & \multicolumn{2}{c}{\textbf{PubHealth}} & \multicolumn{2}{c}{\textbf{SciTab}} & \multicolumn{2}{c}{\textbf{TabFact}} & \multicolumn{2}{c}{\textbf{MMSci}} & \multicolumn{2}{c}{\textbf{PubHealth}} & \multicolumn{2}{c}{\textbf{SciTab}} & \multicolumn{2}{c}{\textbf{TabFact}} & \multicolumn{2}{c}{\textbf{MMSci}} \\
\cmidrule(lr){3-4} \cmidrule(lr){5-6} \cmidrule(lr){7-8} \cmidrule(lr){9-10} \cmidrule(lr){11-12} \cmidrule(lr){13-14} \cmidrule(lr){15-16} \cmidrule(lr){17-18}
& & Acc & F1 & Acc & F1 & Acc & F1 & Acc & F1 & Acc & F1 & Acc & F1 & Acc & F1 & Acc & F1 \\
\midrule
\multirow{4}{*}{Direct} 
& Baseline & 77.8 & 72.8 & 48.3 & 43.4 & 57.6 & 54.6 & 64.7 & 38.9 & 82.8 & 80.4 & 53.6 & 50.7 & 58.6 & 60.3 & 66.7 & 45.0 \\
& +COPRO & 80.6 & 77.2 & 49.9 & 46.4 & 58.8 & 58.9 & 65.9 & \textbf{45.3} & 82.8 & 80.2 & 51.5 & 48.2 & 54.4 & 59.2 & 65.7 & 44.9 \\
& +MiPROv2 & 80.6 & 79.4 & \textbf{55.0} & \textbf{54.7} & \textbf{63.2} & \textbf{64.1} & \textbf{66.9} & 45.0 & 82.8 & 80.3 & 55.9 & 54.6 & 59.2 & 61.6 & \textbf{67.4} & 46.0 \\
& +SIMBA & \textbf{81.7} & \textbf{79.5} & 54.3 & 52.8 & 59.2 & 60.1 & 64.5 & 45.0 & \textbf{85.6} & \textbf{83.7} & \textbf{60.6} & \textbf{60.6} & \textbf{62.9} & \textbf{62.9} & 67.3 & \textbf{47.4} \\
\midrule
\multirow{4}{*}{CoT} 
& Baseline & 87.8 & 86.4 & 54.3 & 52.3 & 75.5 & 77.7 & 79.3 & 54.6 & 87.8 & 86.9 & 62.2 & 61.9 & 78.3 & 80.8 & 82.9 & 58.9 \\
& +COPRO & 87.8 & 86.5 & 57.3 & 56.4 & 74.5 & 76.6 & 79.8 & 54.8 & \textbf{89.4} & \textbf{88.7} & 61.5 & 61.3 & 78.4 & 81.6 & 84.6 & 59.8 \\
& +MiPROv2 & 87.2 & 85.6 & 58.3 & 57.8 & \textbf{80.1} & \textbf{82.2} & \textbf{84.7} & \textbf{60.5} & 88.9 & 87.8 & \textbf{64.8} & \textbf{64.4} & \textbf{81.4} & \textbf{83.4} & \textbf{85.8} & \textbf{62.5} \\
& +SIMBA & \textbf{89.4} & \textbf{88.8} & \textbf{60.1} & \textbf{59.6} & 77.6 & 79.3 & 83.2 & 57.7 & 88.9 & 87.6 & 63.6 & 63.8 & 75.8 & 79.1 & 81.9 & 59.0 \\
\midrule
\multirow{4}{*}{ReAct} 
& Baseline & 83.9 & 82.9 & 49.2 & 48.7 & 64.9 & 72.8 & 79.9 & 57.5 & 87.8 & 86.8 & 52.9 & 52.9 & 76.3 & 80.8 & 82.9 & 58.7 \\
& +COPRO & \textbf{87.2} & \textbf{86.5} & \textbf{58.3} & \textbf{57.1} & 77.1 & 79.4 & 84.7 & \textbf{61.0} & 85.0 & 83.6 & 48.0 & 47.9 & 72.9 & 78.4 & 69.3 & 52.5 \\
& +MiPROv2 & 84.4 & 83.5 & 49.0 & 48.7 & 64.6 & 72.5 & 79.9 & 57.4 & 89.4 & 88.8 & \textbf{63.6} & \textbf{63.2} & 82.9 & 84.4 & \textbf{86.5} & 62.5 \\
& +SIMBA & 86.7 & 85.7 & 53.4 & 51.0 & \textbf{79.8} & \textbf{81.1} & \textbf{84.8} & 59.2 & \textbf{90.0} & \textbf{89.3} & 60.4 & 58.9 & \textbf{84.0} & \textbf{85.0} & 85.8 & \textbf{62.6} \\
\midrule
\multirow{4}{*}{CodeAct} 
& Baseline & 86.7 & 86.0 & 51.5 & 49.8 & 64.7 & 72.2 & 83.2 & 57.7 & 87.2 & 86.2 & 55.9 & 56.1 & 73.6 & 78.7 & 85.8 & 61.3 \\
& +COPRO & \textbf{89.4} & \textbf{88.9} & 54.3 & 53.4 & 67.0 & 74.4 & \textbf{85.0} & 61.1 & 88.9 & 87.9 & \textbf{59.2} & \textbf{59.5} & 79.0 & 81.5 & 84.4 & 61.1 \\
& +MiPROv2 & 88.3 & 87.6 & 49.9 & 48.6 & \textbf{78.9} & \textbf{81.6} & 84.7 & 59.0 & 85.6 & 84.8 & 55.5 & 56.0 & 81.3 & 83.6 & 86.5  & 62.8 \\
& +SIMBA & 85.0 & 84.0 & \textbf{55.2} & \textbf{54.8} & 77.5 & 79.9 & 83.4 & \textbf{61.7} & \textbf{89.4} & \textbf{88.5} & 58.3 & 56.7 & \textbf{83.1} & \textbf{84.4} & \textbf{87.6} & \textbf{65.6} \\
\bottomrule
\end{tabular}
}
\caption{Results of Gemma3-12B and Gemma3-27B on test sets. \textbf{Bold} is best performance per method and dataset. \xt{TODO: Add evaluation of ReActable with Gemma models} \xt{If you have to cut space: You can put Gemma3 table to appendix; then briefly mention Gemma3 underperforms Qwen3 and Gemma3-27B also follows similar trends in the main text}}
\label{tab:gemma3_combined_results}
\end{table*}

\begin{table*}[ht!]
\centering
\resizebox{\textwidth}{!}{
\begin{tabular}{llcccccccccccccccc}
\toprule
& & \multicolumn{8}{c}{\textbf{GPT-4o-mini}} & \multicolumn{8}{c}{\textbf{GPT-4o}} \\
\cmidrule(lr){3-10} \cmidrule(lr){11-18}
\textbf{Module} & \textbf{Optimizer} & \multicolumn{2}{c}{\textbf{PubHealth}} & \multicolumn{2}{c}{\textbf{SciTab}} & \multicolumn{2}{c}{\textbf{TabFact-mini}} & \multicolumn{2}{c}{\textbf{MMSci}} & \multicolumn{2}{c}{\textbf{PubHealth}} & \multicolumn{2}{c}{\textbf{SciTab}} & \multicolumn{2}{c}{\textbf{TabFact-mini}} & \multicolumn{2}{c}{\textbf{MMSci}} \\
\cmidrule(lr){3-4} \cmidrule(lr){5-6} \cmidrule(lr){7-8} \cmidrule(lr){9-10} \cmidrule(lr){11-12} \cmidrule(lr){13-14} \cmidrule(lr){15-16} \cmidrule(lr){17-18}
& & Acc & F1 & Acc & F1 & Acc & F1 & Acc & F1 & Acc & F1 & Acc & F1 & Acc & F1 & Acc & F1 \\
\midrule
ReActable & & 73.9 & 73.4 & 52.7 & 52.4 & 85.5 & 85.5 & 64.3 & 48.2 & 83.3 & 82.6 & 67.6 & 67.6 & 89.8 & 89.7 & 85.4 & 62.8 \\
\midrule
\multirow{4}{*}{Direct} 
& Baseline & 85.6 & 85.3 & 58.3 & 58.4 & 65.0 & \textbf{66.7} & 70.0 & 51.2 & \textbf{90.6} & \textbf{89.8} & 65.0 & 65.0 & 73.2 & 74.8 & 82.1 & 60.1 \\
& +COPRO & \textbf{86.7} & \textbf{87.1} & \textbf{61.1} & \textbf{61.0} & \textbf{65.2} & 65.7 & 70.8 & 51.8 & 89.4 & 88.9 & 64.8 & 64.7 & 76.0 & 77.0 & \textbf{84.7} & 61.2 \\
& +MiPROv2 & 85.0 & 85.2 & 60.1 & 59.8 & 63.5 & 63.9 & \textbf{71.6} & \textbf{52.9} & 90.0 & 89.1 & 65.0 & 65.1 & 74.5 & 76.0 & 84.4 & \textbf{61.7} \\
& +SIMBA & 86.1 & 85.6 & 57.1 & 56.6 & 60.5 & 64.0 & 69.7 & 51.1 & 89.4 & 88.5 & \textbf{65.3} & \textbf{65.2} & \textbf{76.5} & \textbf{77.3} & 82.8 & 59.8 \\
\midrule
\multirow{4}{*}{CoT} 
& Baseline & \textbf{90.6} & \textbf{90.1} & 62.9 & 63.0 & 79.8 & 82.4 & 83.0 & 58.9 & 87.8 & 87.2 & 69.2 & 69.1 & 87.8 & 89.6 & 87.7 & 63.7 \\
& +COPRO & 90.0 & 89.6 & 61.8 & 61.7 & 81.0 & 82.7 & 83.6 & 61.3 & 87.8 & 87.5 & 69.7 & 69.6 & 88.0 & 89.8 & \textbf{88.4} & 65.0 \\
& +MiPROv2 & 89.4 & 88.9 & \textbf{64.8} & \textbf{64.8} & \textbf{81.2} & \textbf{83.0} & 84.4 & 60.9 & 89.4 & 88.9 & \textbf{70.6} & \textbf{70.5} & 88.5 & 89.9 & 88.3 & \textbf{65.8} \\
& +SIMBA & 90.0 & 89.5 & 64.3 & 64.3 & 78.8 & 81.1 & \textbf{84.5} & \textbf{62.2} & \textbf{90.0} & \textbf{89.8} & \textbf{70.6} & \textbf{70.5} & \textbf{90.2} & \textbf{91.4} & 87.9 & 64.3 \\
\midrule
\multirow{4}{*}{ReAct} 
& Baseline & 87.8 & 87.3 & 55.0 & 53.1 & 84.8 & 85.4 & 84.4 & 61.2 & 88.3 & 87.3 & 64.1 & 62.8 & 90.0 & 90.3 & \textbf{89.5} & 66.2 \\
& +COPRO & 89.4 & 88.9 & 59.4 & 58.4 & 82.8 & 83.7 & \textbf{85.7} & 61.8 & \textbf{89.4} & \textbf{88.9} & 67.8 & 67.3 & 90.2 & 91.0 & 88.7 & 67.0 \\
& +MiPROv2 & 90.0 & 89.6 & \textbf{60.1} & \textbf{60.0} & 82.5 & 83.2 & 84.5 & 60.3 & \textbf{89.4} & 88.4 & 66.2 & 65.6 & 90.8 & 91.4 & 88.5 & \textbf{67.6} \\
& +SIMBA & \textbf{91.7} & \textbf{91.1} & \textbf{60.1} & 59.9 & \textbf{84.8} & \textbf{86.1} & 84.0 & \textbf{62.2} & 88.3 & 87.6 & \textbf{68.3} & \textbf{68.3} & \textbf{91.0} & \textbf{92.3} & 88.1 & 64.0 \\
\midrule
\multirow{4}{*}{CodeAct} 
& Baseline & \textbf{84.4} & \textbf{83.7} & \textbf{59.0} & \textbf{58.8} & 82.5 & 83.9 & 84.5 & 60.4 & 87.2 & 86.7 & 63.4 & 62.3 & 90.2 & 90.8 & 89.3 & \textbf{65.4} \\
& +COPRO & \textbf{84.4} & 82.8 & 53.4 & 52.2 & 83.5 & 84.7 & \textbf{85.4} & \textbf{61.3} & 89.4 & 89.0 & 62.9 & 60.7 & 90.5 & 91.4 & \textbf{89.7} & 64.9 \\
& +MiPROv2 & 80.6 & 77.9 & 52.2 & 48.9 & \textbf{85.2} & \textbf{86.6} & 82.9 & 57.8 & \textbf{91.1} & \textbf{90.6} & \textbf{65.0} & \textbf{63.9} & \textbf{91.2} & \textbf{91.7} & 89.2 & 62.6 \\
& +SIMBA & \textbf{84.4} & 83.0 & 55.7 & 54.9 & 81.5 & 83.0 & 84.1 & 58.6 & 88.3 & 87.6 & 61.1 & 60.5 & 90.0 & 91.4 & 89.0 & \textbf{65.4} \\
\bottomrule
\end{tabular}
}
\caption{Results of GPT-4o-mini and GPT-4o on test sets. \textbf{Bold} is best performance per method and dataset.}
\label{tab:gpt4_combined_results_tabfact_mini}
\end{table*}

%
%





\subsection{Effects of Optimizing Instructions on Table Reasoning}
\label{sec:analysis_baseline_cot}

\Cref{tab:qwen3_combined_results,tab:gemma3_combined_results} provide evidence that prompt optimization improves table reasoning performance in both the direct and CoT settings. To analyze this in more detail, \Cref{tab:confusion_direct,tab:confusion_cot} present confusion matrices comparing the differences between SIMBA optimization and the baseline.

\begin{table}[t]
\centering
\renewcommand{\arraystretch}{1.2} 
\setlength{\tabcolsep}{6pt} 
\resizebox{0.85\linewidth}{!}{
\begin{tabular}{@{}ccccc@{}}
\toprule
\multicolumn{2}{c}{\textbf{Direct}} & \multicolumn{2}{c}{\textbf{Gold}} \\ 
\cmidrule(l){3-4}
\multicolumn{2}{c}{(Base$\rightarrow$SIMBA)} & \textbf{Support} & \textbf{Refute} \\ 
\midrule
\multirow{3}{*}{\rotatebox[origin=c]{90}{\textbf{Predict}}} 
  & \textbf{Support} & 3,870\textcolor{red}{$\rightarrow$}2,941 & 2,430\textcolor{blue}{$\rightarrow$}1,270 \\
  & \textbf{Refute}  & \;\; 311\textcolor{red}{$\rightarrow$}1,242   & 1,648\textcolor{blue}{$\rightarrow$}2,880 \\
  & \makecell{\textbf{Not}\\\textbf{Enough Info}} & 125\textcolor{blue}{$\rightarrow$}123     & 225\textcolor{blue}{$\rightarrow$}153 \\
\bottomrule
\end{tabular}
}
\caption{Confusion matrices before and after prompt optimization for Direct prediction with the Qwen3-32B model on TabFact. \textcolor{blue}{$\rightarrow$} and \textcolor{red}{$\rightarrow$} indicate positive and negative changes respectively.}
\label{tab:confusion_direct}
\end{table}

\begin{table}[t]
\centering
\renewcommand{\arraystretch}{1.2} 
\setlength{\tabcolsep}{6pt} 
\resizebox{0.85\linewidth}{!}{
\begin{tabular}{@{}ccccc@{}}
\toprule
\multicolumn{2}{c}{\textbf{CoT}} & \multicolumn{2}{c}{\textbf{Gold}} \\ 
\cmidrule(l){3-4}
\multicolumn{2}{c}{(Base$\rightarrow$SIMBA)} & \textbf{Support} & \textbf{Refute} \\ 
\midrule
\multirow{3}{*}{\rotatebox[origin=c]{90}{\textbf{Predict}}} 
  & \textbf{Support} & 3,643\textcolor{blue}{$\rightarrow$}3,648 & 429\textcolor{blue}{$\rightarrow$}403 \\
  & \textbf{Refute}  & 483\textcolor{red}{$\rightarrow$}499   & 3,628\textcolor{blue}{$\rightarrow$}3,691 \\
  & \makecell{\textbf{Not}\\\textbf{Enough Info}} & 180\textcolor{blue}{$\rightarrow$}159     & 246\textcolor{blue}{$\rightarrow$}209 \\
\bottomrule
\end{tabular}
}
\caption{Confusion matrices before and after prompt optimization for CoT with the Qwen3-32B model on TabFact. \textcolor{blue}{$\rightarrow$} and \textcolor{red}{$\rightarrow$} indicate positive and negative changes respectively.}
\label{tab:confusion_cot}
\end{table}

The most salient pattern observed from \Cref{tab:confusion_direct,tab:confusion_cot} is that optimization increases the proportion of refute predictions, indicating that it leads to more conservative predictions overall. To further explain this effect, we analyzed the optimized instructions for both the direct and CoT modules.

\begin{table*}[]
\small
\resizebox{\linewidth}{!}{
\begin{tabular}{ll}
\toprule
\textbf{Module} & \multicolumn{1}{c}{\textbf{Instruction}} \\
\midrule
Baseline & Verify the given claim against the provided table data. \\
\midrule
Direct (SIMBA) & \begin{tabular}[c]{@{}l@{}}Verify the given claim against the provided table data.\textbackslash{}n\textbackslash{}nIf the module receives a claim that specifies a particular treatment \\ as the 'first' or 'second' alternative for a given condition, it should carefully cross-check the table data to ensure the claim's \\ order of alternatives matches the table. {[}...{]} \underline{If the table contradicts the claim's order, the module should return 'refutes.'}{[}...{]} \\ Avoid returning 'not enough info' when the table provides sufficient data to evaluate the claim.\end{tabular}\\
\midrule
CoT (SIMBA) & \begin{tabular}[c]{@{}l@{}}Verify the given claim against the provided table data.\textbackslash{}n\textbackslash{}nIf the claim refers to the effectiveness or performance of a method in a \\ specific stage, and the table includes evaluation metrics (e.g., accuracy, percentage) for that stage, then \underline{the module should focus} \\ \underline{on comparing the values in the table to determine if the claim is supported or refuted.} Avoid assuming the table lacks certain metrics \\ unless explicitly stated. {[}...{]}\end{tabular}\\
\bottomrule
\end{tabular}
}
\caption{Baseline and optimized instructions for the Qwen3-32B model using the SIMBA optimizer. The prompt parts that exhibit the characteristics analyzed in \Cref{tab:ordinal_terms_analysis,tab:comp_terms_analysis} are underlined. Some parts of the instructions have been omitted for clarity. \xt{Increase the font size}}
\label{tab:inst_example}
\end{table*}

\begin{table}[t]
\centering
\renewcommand{\arraystretch}{1.2}
\setlength{\tabcolsep}{8pt}
\resizebox{\linewidth}{!}{
\begin{tabular}{lcccc}
\toprule
\textbf{Metric} & \multicolumn{2}{c}{\textbf{With Ordinal Terms}} & \multicolumn{2}{c}{\textbf{Without Ordinal Terms}} \\
\cmidrule(lr){2-3} \cmidrule(lr){4-5}
 & Baseline & SIMBA & Baseline & SIMBA \\
\midrule
Instances & \multicolumn{2}{c}{1,499} & \multicolumn{2}{c}{7,110} \\
\midrule
\multicolumn{5}{l}{\textbf{Accuracy by Gold Label}} \\
\midrule
All (overall) & 63.7 & \textbf{69.0} & 64.2 & \textbf{67.3} \\
Refute        & 39.0 & \textbf{70.6} & 38.1 & \textbf{66.1} \\
Support       & \textbf{90.2} & 67.4 & \textbf{89.8} & 67.4 \\
\bottomrule
\end{tabular}
}
\caption{Comparison of Direct prediction with Qwen3-32B before and after optimization (SIMBA) on instances containing ordinal terms \textit{in the claim} (\Cref{ssec:appendix_ordinal_terms}) from TabFact. \xt{Increase the font size}
}
\label{tab:ordinal_terms_analysis}
\end{table}

For direct prompting,
we observe 
that the optimized instructions are often tuned towards ordinal information. For Qwen3-32B optimized with SIMBA, the resulting instruction (\Cref{tab:inst_example}) includes an additional prompt encouraging attention to the order of elements in the claim (e.g., \textit{“If the table contradicts the claim's order, the module should return ‘refutes.’”}). To validate this effect, we defined a set of ordinal terms (see \Cref{ssec:appendix_ordinal_terms}) and analyzed the effectiveness of the optimized instruction depending on whether the claim contained ordinal terms (\Cref{tab:ordinal_terms_analysis}). Indeed, when the gold label is `refute', the optimized instruction shows a notable 4.5\% improvement (from 66.1\% to 70.6\%) for claims containing ordinal terms compared to those without.

\begin{table}[t]
\centering
\resizebox{\linewidth}{!}{
\begin{tabular}{lccc}
\toprule
\textbf{Metric} & \textbf{Baseline} & \textbf{SIMBA} & \textbf{Overlap} \\
\midrule
Instances w/ comparative terms & 1,882 & \textbf{2,016} & 2,382 \\
\midrule
\multicolumn{4}{l}{\textbf{Accuracy on Overlapping Instances by Gold Label}} \\
\midrule
All (overall)  & 83.0 & \textbf{84.1} & -- \\
Refute         & 83.7 & \textbf{86.1} & -- \\
Support        & \textbf{82.2} & 81.8 & -- \\
\bottomrule
\end{tabular}
}
\caption{Comparison of CoT with  Qwen3-32B before and after optimization (SIMBA) on instances containing comparative terms \textit{in the reasoning} (\Cref{ssec:appendix_comp_terms}) from TabFact.
\xt{Increase the font size}
}
\label{tab:comp_terms_analysis}
\end{table}

For CoT, the instruction (\Cref{tab:inst_example}) optimized by SIMBA with the Qwen3-32B model is  particularly specialized to numerical comparisons (e.g., \textit{``the module should focus on comparing the values in the table to determine if the claim is supported or refuted''}). Following a similar procedure as above, we defined a set of comparative terms (\Cref{ssec:appendix_comp_terms}) and analyzed whether the optimized instruction prompts more explicit comparison behavior during reasoning, and how this affects performance (\Cref{tab:comp_terms_analysis}). Again, when the gold label is `refute', the optimized instruction yields a 2.4\% improvement (from 83.7\% to 86.1\%).

\Cref{tab:ordinal_terms_analysis,tab:comp_terms_analysis} validate that the optimization process encourages the model to focus on specific aspects crucial for table reasoning (element order and numerical comparison for Direct prediction and CoT respectively)
and that these behaviors are effectively reflected in performance improvements.

\subsection{Analysis of Tool Use Behavior}
We investigate the tool use behavior of ReAct with Qwen3 models on the SciTab test data before and after instruction optimization. 
For the 32B model, COPRO increases the tool calling frequency from 90\% to 93\%, but leads to more error rates (successful execution is reduced from 42\% to 12\%).
Both MiPROv2 and SIMBA guide the model towards a more direct reasoning path by reducing tool usage frequency from 90\% to 70.6\% and 35.4\% respectively, while increasing the average length of reasoning in the trajectory by 16\% and 33\%.
This implies that MiPROv2 and SIMBA encourage direct table interpretation rather than relying heavily on tool-based verification and only invoke tool calls when necessary.
MiPROv2 leads to more tool execution errors after optimization, while SIMBA maintains a similar level of error rate compared to the baseline.

For the smaller 8B model, COPRO and MiPROv2 dramatically reduce tool usage frequency from 67\% to around 40\%, while SIMBA maintains a similar level of tool use frequency as the baseline with a slightly improved successful execution rate (40.8\% to 41.2\%).

\begin{table}[t]
    \centering
    \small
    \begin{tabular}{lrr}
        \toprule
        {\bf Outcome} & {\bf Qwen3-32B} & {\bf Qwen3-8B} \\
        \midrule
        Both correct & 33 & 22 \\
        CoT only correct & 2 & 5 \\
        ReAct only correct & 6 & 8 \\
        Both wrong & 9 & 15 \\
        \bottomrule
    \end{tabular}
    \caption{Comparison of CoT and ReAct on 50 random SciTab test claims.}
    \label{tab:cot-react-50}
\end{table}

\subsection{Error analysis of ReAct and CodeAct \xt{! this section is updated}}

%
To assess the validity of ReAct and CodeAct results, we sample 20 error cases from each setting\xt{TODO: sample 40 from each of ReAct and CodeAct} with Qwen3-32B on the SciTab test set and conduct a manual error analysis. We define a high-level error taxonomy consisting of three categories: data-related, code-related, and reasoning errors, and analyze which category or categories each erroneous instance belongs to, noting that a single instance can fall into more than one category. 

\xt{TODO: Add statistics of ReAct errors in Giwon's table; I got reasoning:8, data:7, code:5}
The main sources of errors in ReAct are issues with reasoning (45\%) and data-related errors (35 \%).
ReAct often focuses on part of the statement in a complex claim, neglecting other relevant data in tool planning or reasoning steps.
Also, tool traces with execution errors can lead to illogical reasoning that is inconsistent with the provided table data.
Data-related issues include underspecified claims and noisy table structure, which makes it difficult for ReAct to construct correct SQL queries.



\begin{table}[t!]
\small
\centering
\begin{tabular}{lrr}
\toprule
\textbf{Taxonomy}                      &
\textbf{ReAct}  &
\textbf{CodeAct} \\
\midrule
Data-related Errors  & 7 (35\%)  & 14 (70.0\%)                 \\
Code-related Errors & 5 (25\%) & 6 (30.0\%)                 \\
Reasoning Errors    & 9 (45\%) & 3  (15.0\%)                 \\
\bottomrule
\end{tabular}
\caption{Number of error instances of each type found in ReAct and CodeAct predictions with Qwen3-32B on SciTab. }
    \label{tab:codex_error}
\end{table}

For CodeAct, the error analysis in Table \ref{tab:codex_error} shows that the largest share of errors (70\%) stems from issues intrinsic to the SciTab dataset, including subjective or underspecified claims (e.g. undefined notions of “significance”), claims requiring information not present in the table, or incorrect gold labels. Code-related errors account for over 30\% of cases; among 6 such errors, 4 arise from code execution failures due primarily to syntactic issues, indicating that more robust coding agents could substantially reduce this error type. Reasoning errors constitute around 15\% of the total and are claim-related, typically involving claim misinterpretation or failure to recognize that the table lacks the required information, where a "not enough information" label should have been predicted. Notably, even when code execution failed, the model occasionally compensated during the reasoning phase by approximating the intended computation through inspection of the generated code or table contents.

\subsection{Comparison of CoT with ReAct}
To understand the performance differences between CoT reasoning and tool-augmented agents, we conduct a detailed analysis of 50 instances from the SciTab test set and evaluate Qwen3-8B and Qwen3-32B predictions after optimizing instructions with MiPROv2.
As shown in Table~\ref{tab:cot-react-50}, ReAct consistently outperforms CoT across both Qwen3-32B (78\% vs.~70\%) and Qwen3-8B (60\% vs.~54\%) on this subset.
A closer inspection of these 50 claims reveals complementary failure modes.

The primary advantage of ReAct lies in its iterative verification process, which is effective for verifying complex or multi-clause claims that involve multiple quantitative checks, such as ``G2S approaches outperform the S2S baseline''. 
Even though some tool calls may fail with execution errors, the agent can re-examine the table evidence and switch to a different strategy using language-based reasoning in the following step. \xt{why ReAct is better for multi-clause claim?}
This stands in contrast to CoT's reasoning process, which often fails to aggregate all sub-claims before reaching a final verdict.
In addition, CoT often misinterprets metric directionality, such as treating lower error rates (e.g., ADDED/MISS values) as worse performance. 
For Qwen3-32B, four of the six ReAct-only successes are on claims that CoT incorrectly labels as refutes due to misinterpreting metric directionality. The other are cases where CoT abstains from deciding on multi-clause improvements (e.g. "outperforms baselines by 10.5 F1").
The 8B model exhibits a similar pattern but with a stronger tendency to abstain from making decisions.


However, ReAct's reliance on tool-based verification can hinder its performance on simple and straightforward claims, where direct table interpretation is enough for making the decision, such as "BI and IS individually outperform the oracle".
For the Qwen3-8B model, every instance where CoT succeeded but ReAct failed is due to tool execution errors.
The agent is prone to violating the tool schema and representing table data in the wrong format when making tool calls.
The resulting execution errors prevent the verifier from revisiting the claim, causing the reasoning process to terminate with either a \textit{not enough info} or \textit{refutes} label.
Overall, this comparison confirms that tool planning encourages more systematic evidence checking, but the benefits only appear when the learned tool interface is properly aligned with the tool executor.

\subsection{Comparison of ReAct with CodeAct}
\begin{table}[th!]
    \centering
    \small
    \begin{tabular}{lrr}
        \toprule
        {\bf Outcome} & {\bf Baseline} & {\bf with SIMBA} \\
        \midrule
        Both correct & 44 & 52 \\
        ReAct only correct & 17 & 12 \\
        CodeAct only correct & 7 & 13 \\
        Both wrong & 32 & 23 \\
        \bottomrule
    \end{tabular}
    \caption{Comparison of ReAct and CodeAct with Qwen3-32B on 100 random SciTab test instances}
    \label{tab:react-codeact-scitab-100}
\end{table}

We analyze 100 random instances from SciTab test data to study the differences between ReAct and CodeAct based on Qwen3-32B predictions.
Distribution of the correctness before and after SIMBA optimization is shown in \Cref{tab:react-codeact-scitab-100}.

Among 17 ReAct-only correct instances in the baseline experiments, we observe that CodeAct often struggles with parsing table data as well as handling numerical aggregation and comparison. 
Unlike ReAct that uses SQL tool to extract relevant table data and solves math problems in the following reasoning steps, CodeAct relies on python codes for both table parsing and math reasoning.
The generated python code often introduces a table representation or directly assigns relevant table cells to variables, which is prone to mistakes. 
Also, CodeAct tends to be overly strict when making numerical comparisons and may fail to consider all the relevant entries.
After SIMBA optimization, CodeAct gives correct predictions on 11 out of 17 error instances.
We find SIMBA is particularly effective for improving table parsing and simple numerical comparison by introducing heuristic rules.
However, CodeAct still shows limited reasoning ability after optimization in validating complex claims, especially demonstrating weak performance when the claim contains ambiguous descriptions or requires multi-row comparisons. 

\section{Conclusion}
Recent research on tabular fact checking has investigated tool-augmented and agentic approaches, yet it remains unclear whether external tools and program-aided reasoning provide consistent advantages over language-based reasoning for latest LLMs.
In this work, we present the first comparative analysis of various prompting techniques for tabular fact checking and examine the impact of instruction optimization on reasoning and tool use behavior across two model families. 
Our analysis reveals that MiPROv2 optimizer yields substantial gains for CoT reasoning, while SIMBA proves effective in refining the instructions of ReAct and CodeAct agents.
Both MiPROv2 and SIMBA encourage more direct reasoning paths and help reduce unnecessary tool calls.
Overall, CoT remains a strong choice for tabular fact checking, particularly with smaller models.
Meanwhile, ReAct agents built on larger models can achieve competitive performance, but require more careful instruction optimization.


\section*{Limitations}
While our study provides valuable insights into the impact of instruction optimization on diverse prompting techniques for tabular fact checking, we acknowledge several limitations in our experimental setup.
The scope of our experiments is restricted to two specific model families, Qwen3 and Gemma3, each with two sizes, and the generalizability of our findings to other model architectures or substantially larger models remains an open question.
Our comparative analysis is limited to three representative instruction optimization methods, and a broader survey of latest techniques such as GEPA \cite{agrawal2025gepareflectivepromptevolution} could reveal different model behaviors or performance trade-offs.
Our evaluation of ReAct agent is conducted with a single tool for executing SQL queries, which does not capture the complexities of tool selection in multi-tool scenarios.
Additionally, we do not account for the potential influence of the quality of the initial instruction on the optimization process.
All the experiments are conducted with the same seed instruction in a simple format.



\section*{Acknowledgments}
We thank the anonymous reviewers for the insightful feedback and comments.
Xiaotang Du was partly supported by the UKRI Centre for Doctoral Training in Natural Language Processing, funded by UK Research and Innovation (grant EP/S022481/1) and the University of Edinburgh, School of Informatics.
Pasquale Minervini was partially funded by ELIAI (The Edinburgh Laboratory for Integrated Artificial Intelligence), EPSRC (grant no.\ EP/W002876/1), an industry grant from Cisco, and a donation from Accenture LLP.
Giwon Hong was supported by the ILCC PhD program (School of Informatics Funding Package) at the University of Edinburgh, School of Informatics.
\xt{TODO: Input from collaborators}
This work was supported by the Edinburgh International Data Facility (EIDF) and the Data-Driven Innovation Programme at the University of Edinburgh. 
Some evaluations were performed using resources provided by the Cambridge Service for Data Driven Discovery (CSD3) operated by the University of Cambridge Research Computing Service, provided by Dell EMC and Intel using Tier-2 funding from the Engineering and Physical Sciences Research Council (capital grant EP/T022159/1), and DiRAC funding from the Science and Technology Facilities Council. 

\newpage

\bibliography{custom}

@inproceedings{lu2023scitab,
    title = "{SCITAB}: A Challenging Benchmark for Compositional Reasoning and Claim Verification on Scientific Tables",
    author = "Lu, Xinyuan and Pan, Liangming and Liu, Qian and Nakov, Preslav and Kan, Min-Yen",
    booktitle = "Proceedings of the 2023 Conference on Empirical Methods in Natural Language Processing",
    month = dec,
    year = "2023",
    address = "Singapore",
    publisher = "Association for Computational Linguistics",
    pages = "7787--7813",
    doi = "10.18653/v1/2023.emnlp-main.483",
    url = "https://aclanthology.org/2023.emnlp-main.483",
}

@inproceedings{schlichtkrull2023averitec,
    title = "{AVeriTeC}: A Dataset for Real-world Claim Verification with Evidence from the Web",
    author = "Schlichtkrull, Michael and Guo, Zhijiang and Vlachos, Andreas",
    booktitle = "Thirty-seventh Conference on Neural Information Processing Systems Datasets and Benchmarks Track",
    year = "2023",
    url = "https://arxiv.org/abs/2305.13117"
}

@inproceedings{yao2023react,
    title = "{ReAct}: Synergizing Reasoning and Acting in Language Models",
    author = "Yao, Shunyu and Zhao, Jeffrey and Yu, Dian and Du, Nan and Shafran, Izhak and Narasimhan, Karthik and Cao, Yuan",
    booktitle = "Proceedings of the Eleventh International Conference on Learning Representations",
    year = "2023",
    url = "https://openreview.net/forum?id=WE_vluYUL-X",
    publisher = "ICLR"
}

@inproceedings{khattab2023dspy,
  author       = {Omar Khattab and
                  Arnav Singhvi and
                  Paridhi Maheshwari and
                  Zhiyuan Zhang and
                  Keshav Santhanam and
                  Sri Vardhamanan and
                  Saiful Haq and
                  Ashutosh Sharma and
                  Thomas T. Joshi and
                  Hanna Moazam and
                  Heather Miller and
                  Matei Zaharia and
                  Christopher Potts},
  title        = {DSPy: Compiling Declarative Language Model Calls into State-of-the-Art
                  Pipelines},
  booktitle    = {{ICLR}},
  publisher    = {OpenReview.net},
  year         = {2024}
}

@article{zhao2024findver,
    title = "{FinDVer}: Explainable Claim Verification over Long and Hybrid-Content Financial Documents",
    author = "Zhao, Yilun and Long, Yitao and Jiang, Yuru and Wang, Chengye and Chen, Weiyuan and Liu, Hongjun and Zhang, Yiming and Tang, Xiangru and Zhao, Chen and Cohan, Arman",
    year = "2024",
    eprint = "2411.05764",
    archivePrefix = "arXiv",
    primaryClass = "cs.CL",
    url = "https://arxiv.org/abs/2411.05764"
}

@inproceedings{aly-etal-2021-fact,
    title = "The Fact Extraction and {VER}ification Over Unstructured and Structured information ({FEVEROUS}) Shared Task",
    author = "Aly, Rami  and
      Guo, Zhijiang  and
      Schlichtkrull, Michael Sejr  and
      Thorne, James  and
      Vlachos, Andreas  and
      Christodoulopoulos, Christos  and
      Cocarascu, Oana  and
      Mittal, Arpit",
    editor = "Aly, Rami  and
      Christodoulopoulos, Christos  and
      Cocarascu, Oana  and
      Guo, Zhijiang  and
      Mittal, Arpit  and
      Schlichtkrull, Michael  and
      Thorne, James  and
      Vlachos, Andreas",
    booktitle = "Proceedings of the Fourth Workshop on Fact Extraction and VERification (FEVER)",
    month = nov,
    year = "2021",
    address = "Dominican Republic",
    publisher = "Association for Computational Linguistics",
    url = "https://aclanthology.org/2021.fever-1.1/",
    doi = "10.18653/v1/2021.fever-1.1",
    pages = "1--13",
}

@inproceedings{ChenWCZWLZW20,
  author    = {Wenhu Chen and Hongmin Wang and Jianshu Chen and Yunkai Zhang and Hong Wang and Shiyang Li and Xiyou Zhou and William Yang Wang},
  title     = {{TabFact}: A Large-scale Dataset for Table-based Fact Verification},
  booktitle = {Proceedings of the Eighth International Conference on Learning Representations},
  year      = {2020},
  url       = {https://arxiv.org/abs/1909.02164}
}

@inproceedings{wang2024executable,
      title={Executable Code Actions Elicit Better LLM Agents}, 
      author={Xingyao Wang and Yangyi Chen and Lifan Yuan and Yizhe Zhang and Yunzhu Li and Hao Peng and Heng Ji},
      year={2024},
      eprint={2402.01030},
      booktitle={ICML}
}

@article{qwen3,
    title={Qwen3 Technical Report}, 
    author={An Yang and Anfeng Li and Baosong Yang and Beichen Zhang and Binyuan Hui and Bo Zheng and Bowen Yu and Chang Gao and Chengen Huang and Chenxu Lv and Chujie Zheng and Dayiheng Liu and Fan Zhou and Fei Huang and Feng Hu and Hao Ge and Haoran Wei and Huan Lin and Jialong Tang and Jian Yang and Jianhong Tu and Jianwei Zhang and Jianxin Yang and Jiaxi Yang and Jing Zhou and Jingren Zhou and Junyang Lin and Kai Dang and Keqin Bao and Kexin Yang and Le Yu and Lianghao Deng and Mei Li and Mingfeng Xue and Mingze Li and Pei Zhang and Peng Wang and Qin Zhu and Rui Men and Ruize Gao and Shixuan Liu and Shuang Luo and Tianhao Li and Tianyi Tang and Wenbiao Yin and Xingzhang Ren and Xinyu Wang and Xinyu Zhang and Xuancheng Ren and Yang Fan and Yang Su and Yichang Zhang and Yinger Zhang and Yu Wan and Yuqiong Liu and Zekun Wang and Zeyu Cui and Zhenru Zhang and Zhipeng Zhou and Zihan Qiu},
    journal = {arXiv preprint arXiv:2505.09388},
    year={2025}
}

@misc{gemmateam2025gemma3technicalreport,
      title={Gemma 3 Technical Report}, 
      author={Gemma Team and Aishwarya Kamath and Johan Ferret and Shreya Pathak and Nino Vieillard and Ramona Merhej and Sarah Perrin and Tatiana Matejovicova and Alexandre Ramé and Morgane Rivière and Louis Rouillard and Thomas Mesnard and Geoffrey Cideron and Jean-bastien Grill and Sabela Ramos and Edouard Yvinec and Michelle Casbon and Etienne Pot and Ivo Penchev and Gaël Liu and Francesco Visin and Kathleen Kenealy and Lucas Beyer and Xiaohai Zhai and Anton Tsitsulin and Robert Busa-Fekete and Alex Feng and Noveen Sachdeva and Benjamin Coleman and Yi Gao and Basil Mustafa and Iain Barr and Emilio Parisotto and David Tian and Matan Eyal and Colin Cherry and Jan-Thorsten Peter and Danila Sinopalnikov and Surya Bhupatiraju and Rishabh Agarwal and Mehran Kazemi and Dan Malkin and Ravin Kumar and David Vilar and Idan Brusilovsky and Jiaming Luo and Andreas Steiner and Abe Friesen and Abhanshu Sharma and Abheesht Sharma and Adi Mayrav Gilady and Adrian Goedeckemeyer and Alaa Saade and Alex Feng and Alexander Kolesnikov and Alexei Bendebury and Alvin Abdagic and Amit Vadi and András György and André Susano Pinto and Anil Das and Ankur Bapna and Antoine Miech and Antoine Yang and Antonia Paterson and Ashish Shenoy and Ayan Chakrabarti and Bilal Piot and Bo Wu and Bobak Shahriari and Bryce Petrini and Charlie Chen and Charline Le Lan and Christopher A. Choquette-Choo and CJ Carey and Cormac Brick and Daniel Deutsch and Danielle Eisenbud and Dee Cattle and Derek Cheng and Dimitris Paparas and Divyashree Shivakumar Sreepathihalli and Doug Reid and Dustin Tran and Dustin Zelle and Eric Noland and Erwin Huizenga and Eugene Kharitonov and Frederick Liu and Gagik Amirkhanyan and Glenn Cameron and Hadi Hashemi and Hanna Klimczak-Plucińska and Harman Singh and Harsh Mehta and Harshal Tushar Lehri and Hussein Hazimeh and Ian Ballantyne and Idan Szpektor and Ivan Nardini and Jean Pouget-Abadie and Jetha Chan and Joe Stanton and John Wieting and Jonathan Lai and Jordi Orbay and Joseph Fernandez and Josh Newlan and Ju-yeong Ji and Jyotinder Singh and Kat Black and Kathy Yu and Kevin Hui and Kiran Vodrahalli and Klaus Greff and Linhai Qiu and Marcella Valentine and Marina Coelho and Marvin Ritter and Matt Hoffman and Matthew Watson and Mayank Chaturvedi and Michael Moynihan and Min Ma and Nabila Babar and Natasha Noy and Nathan Byrd and Nick Roy and Nikola Momchev and Nilay Chauhan and Noveen Sachdeva and Oskar Bunyan and Pankil Botarda and Paul Caron and Paul Kishan Rubenstein and Phil Culliton and Philipp Schmid and Pier Giuseppe Sessa and Pingmei Xu and Piotr Stanczyk and Pouya Tafti and Rakesh Shivanna and Renjie Wu and Renke Pan and Reza Rokni and Rob Willoughby and Rohith Vallu and Ryan Mullins and Sammy Jerome and Sara Smoot and Sertan Girgin and Shariq Iqbal and Shashir Reddy and Shruti Sheth and Siim Põder and Sijal Bhatnagar and Sindhu Raghuram Panyam and Sivan Eiger and Susan Zhang and Tianqi Liu and Trevor Yacovone and Tyler Liechty and Uday Kalra and Utku Evci and Vedant Misra and Vincent Roseberry and Vlad Feinberg and Vlad Kolesnikov and Woohyun Han and Woosuk Kwon and Xi Chen and Yinlam Chow and Yuvein Zhu and Zichuan Wei and Zoltan Egyed and Victor Cotruta and Minh Giang and Phoebe Kirk and Anand Rao and Kat Black and Nabila Babar and Jessica Lo and Erica Moreira and Luiz Gustavo Martins and Omar Sanseviero and Lucas Gonzalez and Zach Gleicher and Tris Warkentin and Vahab Mirrokni and Evan Senter and Eli Collins and Joelle Barral and Zoubin Ghahramani and Raia Hadsell and Yossi Matias and D. Sculley and Slav Petrov and Noah Fiedel and Noam Shazeer and Oriol Vinyals and Jeff Dean and Demis Hassabis and Koray Kavukcuoglu and Clement Farabet and Elena Buchatskaya and Jean-Baptiste Alayrac and Rohan Anil and Dmitry and Lepikhin and Sebastian Borgeaud and Olivier Bachem and Armand Joulin and Alek Andreev and Cassidy Hardin and Robert Dadashi and Léonard Hussenot},
      year={2025},
      eprint={2503.19786},
      archivePrefix={arXiv},
      primaryClass={cs.CL},
      url={https://arxiv.org/abs/2503.19786}, 
}

@misc{agrawal2025gepareflectivepromptevolution,
      title={GEPA: Reflective Prompt Evolution Can Outperform Reinforcement Learning}, 
      author={Lakshya A Agrawal and Shangyin Tan and Dilara Soylu and Noah Ziems and Rishi Khare and Krista Opsahl-Ong and Arnav Singhvi and Herumb Shandilya and Michael J Ryan and Meng Jiang and Christopher Potts and Koushik Sen and Alexandros G. Dimakis and Ion Stoica and Dan Klein and Matei Zaharia and Omar Khattab},
      year={2025},
      eprint={2507.19457},
      archivePrefix={arXiv},
      primaryClass={cs.CL},
      url={https://arxiv.org/abs/2507.19457}, 
}

@software{liu2024instructor,
  author = {Jason Liu and Contributors},
  title = {Instructor: A library for structured outputs from large language models},
  url = {https://github.com/instructor-ai/instructor},
  year = {2024},
  month = {3}
}

@inproceedings{shi-etal-2020-learn,
    title = "Learn to Combine Linguistic and Symbolic Information for Table-based Fact Verification",
    author = "Shi, Qi  and
      Zhang, Yu  and
      Yin, Qingyu  and
      Liu, Ting",
    editor = "Scott, Donia  and
      Bel, Nuria  and
      Zong, Chengqing",
    booktitle = "Proceedings of the 28th International Conference on Computational Linguistics",
    month = dec,
    year = "2020",
    address = "Barcelona, Spain (Online)",
    publisher = "International Committee on Computational Linguistics",
    url = "https://aclanthology.org/2020.coling-main.466/",
    doi = "10.18653/v1/2020.coling-main.466",
    pages = "5335--5346",
}

@inproceedings{akhtar-etal-2022-pubhealthtab,
    title = "{P}ub{H}ealth{T}ab: {A} Public Health Table-based Dataset for Evidence-based Fact Checking",
    author = "Akhtar, Mubashara  and
      Cocarascu, Oana  and
      Simperl, Elena",
    editor = "Carpuat, Marine  and
      de Marneffe, Marie-Catherine  and
      Meza Ruiz, Ivan Vladimir",
    booktitle = "Findings of the Association for Computational Linguistics: NAACL 2022",
    month = jul,
    year = "2022",
    address = "Seattle, United States",
    publisher = "Association for Computational Linguistics",
    url = "https://aclanthology.org/2022.findings-naacl.1/",
    doi = "10.18653/v1/2022.findings-naacl.1",
    pages = "1--16",
}

@inproceedings{zhang-etal-2020-table,
    title = "Table Fact Verification with Structure-Aware Transformer",
    author = "Zhang, Hongzhi  and
      Wang, Yingyao  and
      Wang, Sirui  and
      Cao, Xuezhi  and
      Zhang, Fuzheng  and
      Wang, Zhongyuan",
    editor = "Webber, Bonnie  and
      Cohn, Trevor  and
      He, Yulan  and
      Liu, Yang",
    booktitle = "Proceedings of the 2020 Conference on Empirical Methods in Natural Language Processing (EMNLP)",
    month = nov,
    year = "2020",
    address = "Online",
    publisher = "Association for Computational Linguistics",
    url = "https://aclanthology.org/2020.emnlp-main.126/",
    doi = "10.18653/v1/2020.emnlp-main.126",
    pages = "1624--1629",
}

@inproceedings{yang-etal-2020-program,
    title = "Program Enhanced Fact Verification with Verbalization and Graph Attention Network",
    author = "Yang, Xiaoyu  and
      Nie, Feng  and
      Feng, Yufei  and
      Liu, Quan  and
      Chen, Zhigang  and
      Zhu, Xiaodan",
    editor = "Webber, Bonnie  and
      Cohn, Trevor  and
      He, Yulan  and
      Liu, Yang",
    booktitle = "Proceedings of the 2020 Conference on Empirical Methods in Natural Language Processing (EMNLP)",
    month = nov,
    year = "2020",
    address = "Online",
    publisher = "Association for Computational Linguistics",
    url = "https://aclanthology.org/2020.emnlp-main.628/",
    doi = "10.18653/v1/2020.emnlp-main.628",
    pages = "7810--7825",
}

@inproceedings{eisenschlos-etal-2020-understanding,
    title = "Understanding tables with intermediate pre-training",
    author = {Eisenschlos, Julian  and
      Krichene, Syrine  and
      M{\"u}ller, Thomas},
    editor = "Cohn, Trevor  and
      He, Yulan  and
      Liu, Yang",
    booktitle = "Findings of the Association for Computational Linguistics: EMNLP 2020",
    month = nov,
    year = "2020",
    address = "Online",
    publisher = "Association for Computational Linguistics",
    url = "https://aclanthology.org/2020.findings-emnlp.27/",
    doi = "10.18653/v1/2020.findings-emnlp.27",
    pages = "281--296",
}

@inproceedings{yang-zhu-2021-exploring-decomposition,
    title = "Exploring Decomposition for Table-based Fact Verification",
    author = "Yang, Xiaoyu  and
      Zhu, Xiaodan",
    editor = "Moens, Marie-Francine  and
      Huang, Xuanjing  and
      Specia, Lucia  and
      Yih, Scott Wen-tau",
    booktitle = "Findings of the Association for Computational Linguistics: EMNLP 2021",
    month = nov,
    year = "2021",
    address = "Punta Cana, Dominican Republic",
    publisher = "Association for Computational Linguistics",
    url = "https://aclanthology.org/2021.findings-emnlp.90/",
    doi = "10.18653/v1/2021.findings-emnlp.90",
    pages = "1045--1052",
}

@inproceedings{zhong-etal-2020-logicalfactchecker,
    title = "{L}ogical{F}act{C}hecker: Leveraging Logical Operations for Fact Checking with Graph Module Network",
    author = "Zhong, Wanjun  and
      Tang, Duyu  and
      Feng, Zhangyin  and
      Duan, Nan  and
      Zhou, Ming  and
      Gong, Ming  and
      Shou, Linjun  and
      Jiang, Daxin  and
      Wang, Jiahai  and
      Yin, Jian",
    editor = "Jurafsky, Dan  and
      Chai, Joyce  and
      Schluter, Natalie  and
      Tetreault, Joel",
    booktitle = "Proceedings of the 58th Annual Meeting of the Association for Computational Linguistics",
    month = jul,
    year = "2020",
    address = "Online",
    publisher = "Association for Computational Linguistics",
    url = "https://aclanthology.org/2020.acl-main.539/",
    doi = "10.18653/v1/2020.acl-main.539",
    pages = "6053--6065",
}

@inproceedings{dong-smith-2021-structural,
    title = "Structural Encoding and Pre-training Matter: Adapting {BERT} for Table-Based Fact Verification",
    author = "Dong, Rui  and
      Smith, David",
    editor = "Merlo, Paola  and
      Tiedemann, Jorg  and
      Tsarfaty, Reut",
    booktitle = "Proceedings of the 16th Conference of the European Chapter of the Association for Computational Linguistics: Main Volume",
    month = apr,
    year = "2021",
    address = "Online",
    publisher = "Association for Computational Linguistics",
    url = "https://aclanthology.org/2021.eacl-main.201/",
    doi = "10.18653/v1/2021.eacl-main.201",
    pages = "2366--2375",
}

@inproceedings{ou-liu-2022-learning,
    title = "Learning to Generate Programs for Table Fact Verification via Structure-Aware Semantic Parsing",
    author = "Ou, Suixin  and
      Liu, Yongmei",
    editor = "Muresan, Smaranda  and
      Nakov, Preslav  and
      Villavicencio, Aline",
    booktitle = "Proceedings of the 60th Annual Meeting of the Association for Computational Linguistics (Volume 1: Long Papers)",
    month = may,
    year = "2022",
    address = "Dublin, Ireland",
    publisher = "Association for Computational Linguistics",
    url = "https://aclanthology.org/2022.acl-long.525/",
    doi = "10.18653/v1/2022.acl-long.525",
    pages = "7624--7638",
}

@inproceedings{Cheng2022BindingLM,
  author       = {Zhoujun Cheng and
                  Tianbao Xie and
                  Peng Shi and
                  Chengzu Li and
                  Rahul Nadkarni and
                  Yushi Hu and
                  Caiming Xiong and
                  Dragomir Radev and
                  Mari Ostendorf and
                  Luke Zettlemoyer and
                  Noah A. Smith and
                  Tao Yu},
  title        = {Binding Language Models in Symbolic Languages},
  booktitle    = {{ICLR}},
  publisher    = {OpenReview.net},
  year         = {2023}
}

@inproceedings{wu-feng-2024-protrix,
    title = "{P}ro{T}rix: Building Models for Planning and Reasoning over Tables with Sentence Context",
    author = "Wu, Zirui  and
      Feng, Yansong",
    editor = "Al-Onaizan, Yaser  and
      Bansal, Mohit  and
      Chen, Yun-Nung",
    booktitle = "Findings of the Association for Computational Linguistics: EMNLP 2024",
    month = nov,
    year = "2024",
    address = "Miami, Florida, USA",
    publisher = "Association for Computational Linguistics",
    url = "https://aclanthology.org/2024.findings-emnlp.253/",
    doi = "10.18653/v1/2024.findings-emnlp.253",
    pages = "4378--4406",
}

@inproceedings{glenn-etal-2024-blendsql,
    title = "{B}lend{SQL}: A Scalable Dialect for Unifying Hybrid Question Answering in Relational Algebra",
    author = "Glenn, Parker  and
      Dakle, Parag  and
      Wang, Liang  and
      Raghavan, Preethi",
    editor = "Ku, Lun-Wei  and
      Martins, Andre  and
      Srikumar, Vivek",
    booktitle = "Findings of the Association for Computational Linguistics: ACL 2024",
    month = aug,
    year = "2024",
    address = "Bangkok, Thailand",
    publisher = "Association for Computational Linguistics",
    url = "https://aclanthology.org/2024.findings-acl.25/",
    doi = "10.18653/v1/2024.findings-acl.25",
    pages = "453--466",
}

@article{aly-vlachos-2024-tabver,
    title = "{T}ab{V}er: Tabular Fact Verification with Natural Logic",
    author = "Aly, Rami  and
      Vlachos, Andreas",
    journal = "Transactions of the Association for Computational Linguistics",
    volume = "12",
    year = "2024",
    address = "Cambridge, MA",
    publisher = "MIT Press",
    url = "https://aclanthology.org/2024.tacl-1.89/",
    doi = "10.1162/tacl_a_00722",
    pages = "1648--1671",
}

@article{zhang2025atomic,
  title={Atomic Reasoning for Scientific Table Claim Verification},
  author={Zhang, Yuji and Wang, Qingyun and Qian, Cheng and Liu, Jiateng and Sun, Chenkai and Zhang, Denghui and Abdelzaher, Tarek and Zhai, Chengxiang and Nakov, Preslav and Ji, Heng},
  journal={arXiv preprint arXiv:2506.06972},
  year={2025}
}

@inproceedings{zhang-etal-2024-tablellama,
    title = "{T}able{L}lama: Towards Open Large Generalist Models for Tables",
    author = "Zhang, Tianshu  and
      Yue, Xiang  and
      Li, Yifei  and
      Sun, Huan",
    editor = "Duh, Kevin  and
      Gomez, Helena  and
      Bethard, Steven",
    booktitle = "Proceedings of the 2024 Conference of the North American Chapter of the Association for Computational Linguistics: Human Language Technologies (Volume 1: Long Papers)",
    month = jun,
    year = "2024",
    address = "Mexico City, Mexico",
    publisher = "Association for Computational Linguistics",
    url = "https://aclanthology.org/2024.naacl-long.335/",
    doi = "10.18653/v1/2024.naacl-long.335",
    pages = "6024--6044",
}

@article{wang2024chain,
  title={Chain-of-table: Evolving tables in the reasoning chain for table understanding},
  author={Wang, Zilong and Zhang, Hao and Li, Chun-Liang and Eisenschlos, Julian Martin and Perot, Vincent and Wang, Zifeng and Miculicich, Lesly and Fujii, Yasuhisa and Shang, Jingbo and Lee, Chen-Yu and others},
  journal={arXiv preprint arXiv:2401.04398},
  year={2024}
}

@inproceedings{lu-etal-2025-tart,
    title = "{TART}: An Open-Source Tool-Augmented Framework for Explainable Table-based Reasoning",
    author = "Lu, Xinyuan  and
      Pan, Liangming  and
      Ma, Yubo  and
      Nakov, Preslav  and
      Kan, Min-Yen",
    editor = "Chiruzzo, Luis  and
      Ritter, Alan  and
      Wang, Lu",
    booktitle = "Findings of the Association for Computational Linguistics: NAACL 2025",
    month = apr,
    year = "2025",
    address = "Albuquerque, New Mexico",
    publisher = "Association for Computational Linguistics",
    url = "https://aclanthology.org/2025.findings-naacl.244/",
    doi = "10.18653/v1/2025.findings-naacl.244",
    pages = "4323--4339",
    ISBN = "979-8-89176-195-7",
}

@inproceedings{zhou-etal-2025-efficient,
    title = "Efficient Multi-Agent Collaboration with Tool Use for Online Planning in Complex Table Question Answering",
    author = "Zhou, Wei  and
      Mesgar, Mohsen  and
      Friedrich, Annemarie  and
      Adel, Heike",
    editor = "Chiruzzo, Luis  and
      Ritter, Alan  and
      Wang, Lu",
    booktitle = "Findings of the Association for Computational Linguistics: NAACL 2025",
    month = apr,
    year = "2025",
    address = "Albuquerque, New Mexico",
    publisher = "Association for Computational Linguistics",
    url = "https://aclanthology.org/2025.findings-naacl.54/",
    doi = "10.18653/v1/2025.findings-naacl.54",
    pages = "945--968",
    ISBN = "979-8-89176-195-7",
}

@article{jiang2025tablemind,
  title={TableMind: An Autonomous Programmatic Agent for Tool-Augmented Table Reasoning},
  author={Jiang, Chuang and Cheng, Mingyue and Tao, Xiaoyu and Mao, Qingyang and Ouyang, Jie and Liu, Qi},
  journal={arXiv preprint arXiv:2509.06278},
  year={2025}
}

@inproceedings{abhyankar-etal-2025-h,
    title = "{H}-{STAR}: {LLM}-driven Hybrid {SQL}-Text Adaptive Reasoning on Tables",
    author = "Abhyankar, Nikhil  and
      Gupta, Vivek  and
      Roth, Dan  and
      Reddy, Chandan K.",
    editor = "Chiruzzo, Luis  and
      Ritter, Alan  and
      Wang, Lu",
    booktitle = "Proceedings of the 2025 Conference of the Nations of the Americas Chapter of the Association for Computational Linguistics: Human Language Technologies (Volume 1: Long Papers)",
    month = apr,
    year = "2025",
    address = "Albuquerque, New Mexico",
    publisher = "Association for Computational Linguistics",
    url = "https://aclanthology.org/2025.naacl-long.445/",
    doi = "10.18653/v1/2025.naacl-long.445",
    pages = "8841--8863",
    ISBN = "979-8-89176-189-6",
}

@inproceedings{wang-etal-2021-semeval,
    title = "{S}em{E}val-2021 Task 9: Fact Verification and Evidence Finding for Tabular Data in Scientific Documents ({SEM}-{TAB}-{FACTS})",
    author = "Wang, Nancy X. R.  and
      Mahajan, Diwakar  and
      Danilevsky, Marina  and
      Rosenthal, Sara",
    editor = "Palmer, Alexis  and
      Schneider, Nathan  and
      Schluter, Natalie  and
      Emerson, Guy  and
      Herbelot, Aurelie  and
      Zhu, Xiaodan",
    booktitle = "Proceedings of the 15th International Workshop on Semantic Evaluation (SemEval-2021)",
    month = aug,
    year = "2021",
    address = "Online",
    publisher = "Association for Computational Linguistics",
    url = "https://aclanthology.org/2021.semeval-1.39/",
    doi = "10.18653/v1/2021.semeval-1.39",
    pages = "317--326",
}

@inproceedings{wang-etal-2025-sciver,
    title = "{S}ci{V}er: Evaluating Foundation Models for Multimodal Scientific Claim Verification",
    author = "Wang, Chengye  and
      Shen, Yifei  and
      Kuang, Zexi  and
      Cohan, Arman  and
      Zhao, Yilun",
    editor = "Che, Wanxiang  and
      Nabende, Joyce  and
      Shutova, Ekaterina  and
      Pilehvar, Mohammad Taher",
    booktitle = "Proceedings of the 63rd Annual Meeting of the Association for Computational Linguistics (Volume 1: Long Papers)",
    month = jul,
    year = "2025",
    address = "Vienna, Austria",
    publisher = "Association for Computational Linguistics",
    url = "https://aclanthology.org/2025.acl-long.420/",
    doi = "10.18653/v1/2025.acl-long.420",
    pages = "8562--8579",
    ISBN = "979-8-89176-251-0",
}

@inproceedings{chan-etal-2024-overview,
    title = "Overview of the Context24 Shared Task on Contextualizing Scientific Claims",
    author = "Chan, Chu Sern Joel  and
      Naik, Aakanksha  and
      Akamatsu, Matthew  and
      Bekele, Hanna  and
      Bransom, Erin  and
      Campbell, Ian  and
      Sparks, Jenna",
    editor = "Ghosal, Tirthankar  and
      Singh, Amanpreet  and
      Waard, Anita  and
      Mayr, Philipp  and
      Naik, Aakanksha  and
      Weller, Orion  and
      Lee, Yoonjoo  and
      Shen, Shannon  and
      Qin, Yanxia",
    booktitle = "Proceedings of the Fourth Workshop on Scholarly Document Processing (SDP 2024)",
    month = aug,
    year = "2024",
    address = "Bangkok, Thailand",
    publisher = "Association for Computational Linguistics",
    url = "https://aclanthology.org/2024.sdp-1.3/",
    pages = "12--21",
}

@article{yang2025does,
  title={Does table source matter? benchmarking and improving multimodal scientific table understanding and reasoning},
  author={Yang, Bohao and Zhang, Yingji and Liu, Dong and Freitas, Andr{\'e} and Lin, Chenghua},
  journal={arXiv preprint arXiv:2501.13042},
  year={2025}
}

@inproceedings{herzig-etal-2020-tapas,
    title = "{T}a{P}as: Weakly Supervised Table Parsing via Pre-training",
    author = {Herzig, Jonathan  and
      Nowak, Pawel Krzysztof  and
      M{\"u}ller, Thomas  and
      Piccinno, Francesco  and
      Eisenschlos, Julian},
    editor = "Jurafsky, Dan  and
      Chai, Joyce  and
      Schluter, Natalie  and
      Tetreault, Joel",
    booktitle = "Proceedings of the 58th Annual Meeting of the Association for Computational Linguistics",
    month = jul,
    year = "2020",
    address = "Online",
    publisher = "Association for Computational Linguistics",
    url = "https://aclanthology.org/2020.acl-main.398/",
    doi = "10.18653/v1/2020.acl-main.398",
    pages = "4320--4333",
}

@inproceedings{Liu2021TAPEXTP,
  author       = {Qian Liu and
                  Bei Chen and
                  Jiaqi Guo and
                  Morteza Ziyadi and
                  Zeqi Lin and
                  Weizhu Chen and
                  Jian{-}Guang Lou},
  title        = {{TAPEX:} Table Pre-training via Learning a Neural {SQL} Executor},
  booktitle    = {{ICLR}},
  publisher    = {OpenReview.net},
  year         = {2022}
}

@article{zhang2024reactable,
  title={ReAcTable: enhancing ReAct for table question answering},
  author={Zhang, Yunjia and Henkel, Jordan and Floratou, Avrilia and Cahoon, Joyce and Deep, Shaleen and Patel, Jignesh M},
  journal={Proceedings of the VLDB Endowment},
  volume={17},
  number={8},
  pages={1981--1994},
  year={2024},
  publisher={VLDB Endowment}
}

@article{wei2022chain,
  title={Chain-of-thought prompting elicits reasoning in large language models},
  author={Wei, Jason and Wang, Xuezhi and Schuurmans, Dale and Bosma, Maarten and Xia, Fei and Chi, Ed and Le, Quoc V and Zhou, Denny and others},
  journal={Advances in neural information processing systems},
  volume={35},
  pages={24824--24837},
  year={2022}
}

@article{bhandari2024exploring,
  author       = {Kushal Raj Bhandari and
                  Sixue Xing and
                  Soham Dan and
                  Jianxi Gao},
  title        = {Exploring the Robustness of Language Models for Tabular Question Answering
                  via Attention Analysis},
  journal      = {Trans. Mach. Learn. Res.},
  volume       = {2025},
  year         = {2025}
}

@article{wu2025tabular,
  title={Tabular Data Understanding with LLMs: A Survey of Recent Advances and Challenges},
  author={Wu, Xiaofeng and Ritter, Alan and Xu, Wei},
  journal={arXiv preprint arXiv:2508.00217},
  year={2025}
}

@inproceedings{webson-pavlick-2022-prompt,
    title = "Do Prompt-Based Models Really Understand the Meaning of Their Prompts?",
    author = "Webson, Albert  and
      Pavlick, Ellie",
    editor = "Carpuat, Marine  and
      de Marneffe, Marie-Catherine  and
      Meza Ruiz, Ivan Vladimir",
    booktitle = "Proceedings of the 2022 Conference of the North American Chapter of the Association for Computational Linguistics: Human Language Technologies",
    month = jul,
    year = "2022",
    address = "Seattle, United States",
    publisher = "Association for Computational Linguistics",
    url = "https://aclanthology.org/2022.naacl-main.167/",
    doi = "10.18653/v1/2022.naacl-main.167",
    pages = "2300--2344",
}

@inproceedings{leidinger-etal-2023-language,
    title = "The language of prompting: What linguistic properties make a prompt successful?",
    author = "Leidinger, Alina  and
      van Rooij, Robert  and
      Shutova, Ekaterina",
    editor = "Bouamor, Houda  and
      Pino, Juan  and
      Bali, Kalika",
    booktitle = "Findings of the Association for Computational Linguistics: EMNLP 2023",
    month = dec,
    year = "2023",
    address = "Singapore",
    publisher = "Association for Computational Linguistics",
    url = "https://aclanthology.org/2023.findings-emnlp.618/",
    doi = "10.18653/v1/2023.findings-emnlp.618",
    pages = "9210--9232",
}

@inproceedings{opsahl-ong-etal-2024-optimizing,
    title = "Optimizing Instructions and Demonstrations for Multi-Stage Language Model Programs",
    author = "Opsahl-Ong, Krista  and
      Ryan, Michael J  and
      Purtell, Josh  and
      Broman, David  and
      Potts, Christopher  and
      Zaharia, Matei  and
      Khattab, Omar",
    editor = "Al-Onaizan, Yaser  and
      Bansal, Mohit  and
      Chen, Yun-Nung",
    booktitle = "Proceedings of the 2024 Conference on Empirical Methods in Natural Language Processing",
    month = nov,
    year = "2024",
    address = "Miami, Florida, USA",
    publisher = "Association for Computational Linguistics",
    url = "https://aclanthology.org/2024.emnlp-main.525/",
    doi = "10.18653/v1/2024.emnlp-main.525",
    pages = "9340--9366"
}

\newpage

\appendix

\section{Implementation Details}
\label{sec:appendix_impl}

\subsection{List of Ordinal Terms}
\label{ssec:appendix_ordinal_terms}

\Cref{tab:ordinal_terms} shows the list of ordinal terms used in the experiments in \Cref{sec:analysis_baseline_cot}. These terms are pre-determined prior to the experiments and used to identify cases where the claim involves ordinal information, such as ordering or ranking relations between the claim and the table values or among the table values themselves.

\subsection{List of Comparative Terms}
\label{ssec:appendix_comp_terms}

\Cref{tab:comp_terms} shows the list of comparative terms used in the experiments in \Cref{sec:analysis_baseline_cot}. These terms are pre-determined prior to the experiments and served as a proxy indicator for whether the model’s reasoning process involves comparisons either between the claim and the table values or among the table values themselves.

\subsection{Hyperparameter setting}
We conduct our analysis with the instruction optimizers from DSPy.
\paragraph{COPRO}
For COPRO optimization, we set the breadth to 6 and depth to 3. The initial temperature for instruction generation is set to 1.2.
\paragraph{MiPROv2}
%
To ensure a consistent comparison among the optimizers, MiPROv2 optimizer is configured to refine only the prompt instructions.
Specifically, we set the number of bootstrapped demonstrations and number of labeled demonstrations to 0 and use the medium optimization mode.
\paragraph{SIMBA}
We set the number of generated candidates per iteration to 6 and the number of iteration steps to 8.
The maximal number of few-shot demonstrations is set to 0 to ensure only the instructions are optimized.
We use a random seed of 0 for sampling candidate programs during optimization.

\subsection{Dataset Details}
All the evaluation datasets (PubHealthTab, SciTab, TabFact, MMSci) used in our experiments are available under the MIT License.
For each dataset, we adopt the original version from the official Github repository.

\subsection{Computation Details}
Experiments with Qwen3-8B, Gemma3-12B and Gemma3-27B were run on a single NVIDIA A100 GPU with 80GB of GPU memory. 
Experiments with Qwen3-32B were run on two NVIDIA A100 GPUs.
The GPU hours depend on the model size, prompting method, optimizer type and size of test dataset.
For example, optimizing the instructions in ReAct or CodeAct with COPRO and SIMBA optimizer on the hybrid train data takes up to 10 hours.
Evaluation on TabFact test data with ReAct or CodeAct framework using a large backbone model (27B/32B) usually takes up to 2 to 3 days.

\subsection{Implementation of ReActable}
The official ReActable implementation is incompatible with GPT-4o models, since the models do not consistently follow the output format defined in the in-context samples, which leads to parsing failure when extracting code blocks from generated content. 
To evaluate ReActable with GPT-4o models, we reproduce the pipeline by enforcing structured output generation using \texttt{instructor} library \cite{liu2024instructor}. 
At each iteration step, the model can choose from three actions, including executing SQL command, executing Python snippet and predicting the final verdict. 
The code is executed on the most recent intermediate table that can be successfully processed. 
We set the maximum number of iterations to 5, use a temperature of 0 and limit the generation to 3,500 tokens per iteration step.
The model is forced to output the final label if it generates repeated code blocks in consecutive iterations or if code execution fails.

We extend the prompt templates by adapting the instructions to our tasks and design system prompts with more detailed instructions to provide guidance on structured output generation, as shown in Figure~\ref{system_prompt_two_class}, \ref{system_prompt_three_class}, \ref{task_prompt_two_class} and \ref{task_prompt_three_class}. We use the same in-context demonstrations across all three-class fact checking experiments, which are built from 5 random instances in SciTab training set. 
We use 3 out of these in-context samples annotated with binary labels in TabFact experiments.

\section{Optimized Instructions}
\label{sec:appendix_inst}

\begin{table*}[t]
\footnotesize
\centering
\setlength{\tabcolsep}{4pt}
\tolerance=1000
\emergencystretch=3em
\begin{tabular}{>{\justifying\arraybackslash}p{0.105\textwidth}>{\justifying\arraybackslash}p{0.835\textwidth}}
\toprule
\textbf{Optimizer} & \multicolumn{1}{c}{\textbf{Optimized Instruction}} \\
\midrule
Baseline & Verify the given claim against the provided table data. \\
\midrule
COPRO & You are a reasoning agent responsible for determining whether a provided \texttt{claim} is supported, inaccurate, or indeterminate based on the data in the provided \texttt{table} and its contextual \texttt{caption}. You will analyze the structure, content, and context of the table, craft an effective SQL query using the \texttt{execute\_sql} tool to extract or verify relevant evidence, and use the retrieved information to make an evidence-based evaluation.\textbackslash{}n\textbackslash{}nFollow the structured approach:\textbackslash{}n1. Begin by thoroughly reading the \texttt{claim}, \texttt{table}, and the \texttt{caption} to understand the context and what the table represents.\textbackslash{}n\textbackslash{}n2. Design an SQL query to extract relevant data directly from the \texttt{table} by executing the \texttt{execute\_sql} tool with a \texttt{table\_name} and \texttt{sql\_query} that targets the claim.\textbackslash{}n3. Examine the query result closely and evaluate it in light of the claim.\textbackslash{}n4. \uline{Reason through your findings step by step, making explicit references to the evidence derived from the query and table.}\textbackslash{}n5. Once you have sufficient information and a conclusive understanding of whether the claim is supported by the data or contradicts it, use the \texttt{finish} tool to provide your final answer: \texttt{supported}, \texttt{not~supported}, or \texttt{insufficient~evidence}.\textbackslash{}n\textbackslash{}n\uline{Your answer should be data-driven and well-supported, with clear reasoning.} \\
\midrule
MiPROv2 & Carefully verify the accuracy of the provided claim using the structured table data. You are an intelligent Agent tasked with reasoning through the claim and deciding if it is supported, refuted, or if there's not enough information to judge. \uline{Use the \texttt{execute\_sql} tool to query the table when necessary for deeper insights} and the \texttt{finish} tool once you are confident in your verification. \uline{Always base your reasoning on the data and avoid making assumptions not directly supported by the table.} \\
\midrule
SIMBA & Verify the given claim against the provided table data.\textbackslash{}n\textbackslash{}nYou are an Agent. In each episode, you will be given the fields \texttt{claim}, \texttt{table}, \texttt{caption} as input. And you can see your past trajectory so far. Your goal is to use one or more of the supplied tools to collect any necessary information for producing \texttt{answer}.\textbackslash{}n\textbackslash{}nTo do this, you will interleave \texttt{next\_thought}, \texttt{next\_tool\_name}, and \texttt{next\_tool\_args} in each turn, and also when finishing the task. After each tool call, you receive a resulting observation, which gets appended to your trajectory.\textbackslash{}n\textbackslash{}nWhen writing \texttt{next\_thought}, you may reason about the current situation and plan for future steps. When selecting the \texttt{next\_tool\_name} and its \texttt{next\_tool\_args}, the tool must be one of:\textbackslash{}n\textbackslash{}n(1) \texttt{execute\_sql} {[}...{]}\textbackslash{}n(2) \texttt{finish} {[}...{]}\textbackslash{}nWhen providing \texttt{next\_tool\_args}, the value inside the field must be in JSON format\textbackslash{}n\textbackslash{}n\uline{If the table data explicitly contains the information needed to verify the claim, the module should immediately call the \texttt{finish} tool without executing unnecessary SQL queries.} Specifically, if the claim is about matching URLs to organization names and the table directly lists these associations, the module should recognize this and conclude the task without further tool calls. {[}...{]} If the claim refers to a specific metric (e.g., accuracy) and a specific setting (e.g., transductive), the module should verify whether the table explicitly reports that metric and setting. \uline{If the table only provides related metrics (e.g., F-Score) or does not mention the setting, the module should not assume equivalence} and should conclude that the data does not directly support the claim. \\
\bottomrule
\end{tabular}
\caption{Seed instruction and the instructions optimized by different optimizers (COPRO, MiPROv2, SIMBA) for ReAct agent with Qwen3-32B model. The underlined instructions highlight key characteristics of each optimizer. Some parts of the instructions have been omitted for clarity.}
\label{tab:optimized_instructions}
\end{table*}


\Cref{tab:optimized_instructions} shows the optimized instructions with different optimizers for Qwen3-32B.
With COPRO optimization, the refined instruction describes a detailed procedural workflow, where ``design an SQL query to extract relevant data'' is included as a mandatory reasoning step.
The agent is forced to use the tool for simple queries, which can lead to a high volume of low-quality or unnecessary tool calls.
With MiPROv2, the optimized instruction only consists of short and high-level guiding principles.
It explicitly mentions "Use the \texttt{execute\_sql} tool when necessary for deeper insights", which allows the agent to decide when to invoke the tool based on claim complexity.
SIMBA optimization results in lengthy instructions by appending specific rules to the seed instruction, including details on how to formulate tool calls and when to avoid constructing complex SQL queries.
By providing explicit heuristics, SIMBA effectively teaches the agent a more sophisticated decision-making strategy that not only reduces unnecessary tool calls but also helps avoid reasoning pitfalls (e.g. the agent should not equate different evaluation metrics if the metric name is not explicitly mentioned in the table). 

\Cref{tab:inst_qwen8b_direct,tab:inst_qwen8b_cot,tab:inst_qwen8b_react,tab:inst_qwen8b_codeact,tab:inst_qwen32b_direct,tab:inst_qwen32b_cot,tab:inst_qwen32b_react,tab:inst_qwen32b_codeact,tab:inst_gemma12b_direct,tab:inst_gemma12b_cot,tab:inst_gemma12b_react,tab:inst_gemma12b_codeact,tab:inst_gemma27b_direct,tab:inst_gemma27b_cot,tab:inst_gemma27b_react,tab:inst_gemma27b_codeact} shows optimized instructions with different optimizers (COPRO, MiPROv2, and SIMBA) for four prompting techniques (direct prompting, CoT, ReAct, and CodeAct) on Qwen3-8B, Qwen3-32B, Gemma3-12B, and Gemma3-27B models.

\section{Ablation Study}
\label{sec:ablation_study}
\subsection{Robustness to Random Seed}
We conduct ablation studies with different random seeds to verify the sensitivity of instruction optimization methods. 
We only adjust the random seeds in MiPROv2 and SIMBA optimizer, since COPRO does not rely on random seeds.
In MiPROv2 and SIMBA, the random seed is used to sample prompting strategies for generating new candidate instructions and to create mini batches of validation data for assessing the quality of proposed instructions.
We evaluate the performance of Qwen3-32B with CoT prompting across three random seeds.
In \Cref{tab:qwen3_cot_seed_ablation}, we observe consistent performance gains in F1 scores on SciTab, TabFact and MMSci test sets with both optimizers.
The variation in metric values is under 2\% in most settings for each optimizer, which indicates that the optimizers are not highly sensitive to varying random seeds.

\begin{table*}[ht!]
\centering
\small
\scalebox{1.0}{
\begin{tabular}{llcccccccccccccccc}
\toprule
& & \multicolumn{2}{c}{\textbf{PubHealth}} & \multicolumn{2}{c}{\textbf{SciTab}} & \multicolumn{2}{c}{\textbf{TabFact}} & \multicolumn{2}{c}{\textbf{MMSci}} \\
\cmidrule(lr){3-4} \cmidrule(lr){5-6} \cmidrule(lr){7-8} \cmidrule(lr){9-10}
\textbf{Random Seed} & \textbf{Optimizer} & Acc & F1 & Acc & F1 & Acc & F1 & Acc & F1 \\
\midrule
\multirow{3}{*}{0} 
& Baseline & 88.3 & 87.6 & 66.4 & 66.4 & 84.5 & 86.6 & 86.5 & 61.6 \\
& +MiPROv2 & 87.8 & 87.3 & 67.4 & 67.2 & 85.9 & 87.8 & 87.4 & 63.6 \\
& +SIMBA & 87.2 & 86.2 & \textbf{68.8} & \textbf{68.6} & 85.2 & 87.1 & 86.4 & 63.1 \\
\midrule
\multirow{2}{*}{9}
& +MiPROv2 & 88.3 & 87.2 & 68.5 & 68.5 & 85.2 & 87.4 & 86.6 & 62.4 \\
& +SIMBA & 88.3 & 87.4 & 67.8 & 67.5 & 86.1 & 87.4 & 87.8 & \textbf{65.1} \\
\midrule
\multirow{2}{*}{42}
& +MiPROv2 & 87.8 & 87.1 & 67.8 & 67.8 & \textbf{86.5} & \textbf{88.3} & 87.3 & \textbf{65.1} \\
& +SIMBA & \textbf{88.9} & \textbf{88.0} & \textbf{68.8} & \textbf{68.6} & \textbf{86.5} & 87.9 & \textbf{88.5} & 64.7 \\
\bottomrule
\end{tabular}
}
\caption{Ablation study on random seeds for Qwen3-32B with CoT. \textbf{Bold} indicates best performance per dataset.}
\label{tab:qwen3_cot_seed_ablation}
\end{table*}

\subsection{Effect of Initial Instruction Quality}
To assess the effect of initial instruction quality on instruction optimization, we optimize Qwen3-32B with CoT prompting using three different seed instructions.
We consider seed instructions of diverse informativeness, including an empty instruction, a paraphrase of the default instruction used in previous analysis, and a more detailed instruction.
\Cref{tab:qwen3_cot_seed_instruction_ablation} demonstrates that different optimizers exhibit varying sensitivity to initial instruction quality.
COPRO benefits from more informative and detailed initial instructions, showing steadily increasing F1 scores on TabFact and MMSci as the seed instruction contains more information.
This is mainly because COPRO proposes new candidate instructions solely based on previous candidates and their evaluation scores. 
A detailed seed instruction can provide more guidance for instruction generation in early iterations.

SIMBA reaches peak performance with concise seed instructions, whereas introducing additional detail in the seed can lead to degradation. 
Since SIMBA appends heuristic rules to the initial instructions, starting from a detailed instruction may restrict the search space, which negatively impacts the generalisability of optimized instructions.
Given semantically equivalent seed instructions, SIMBA is least sensitive to variations in surface form compared with COPRO and MiPROv2. 
The model performance after SIMBA optimization with the paraphrased initial instruction remains similar to the baseline across all test sets.

\begin{table*}[ht!]
\centering
\small
\scalebox{0.8}{
\begin{tabular}{llcccccccc}
\toprule
& & \multicolumn{2}{c}{\textbf{PubHealth}} & \multicolumn{2}{c}{\textbf{SciTab}} & \multicolumn{2}{c}{\textbf{TabFact}} & \multicolumn{2}{c}{\textbf{MMSci}} \\
\cmidrule(lr){3-4} \cmidrule(lr){5-6} \cmidrule(lr){7-8} \cmidrule(lr){9-10}
\textbf{Seed Instruction} & \textbf{Optimizer} & Acc & F1 & Acc & F1 & Acc & F1 & Acc & F1 \\
\midrule
\multirow{4}{*}{\makecell[l]{Verify the given claim against the provided table data. (Default)}} 
& Baseline & 88.3 & 87.6 & 66.4 & 66.4 & 84.5 & 86.6 & 86.5 & 61.6 \\
& +COPRO & 87.2 & 86.1 & 67.4 & 67.3 & 85.5 & 87.6 & 86.7 & 61.5 \\
& +MiPROv2 & 87.2 & 86.5 & \textbf{68.8} & \textbf{68.6} & \textbf{86.9} & \textbf{88.5} & \textbf{87.7} & \textbf{65.4} \\
& +SIMBA & \textbf{90.0} & \textbf{89.6} & \textbf{68.8} & \textbf{68.6} & 85.2 & 87.1 & 87.0 & 64.2 \\
\midrule
\multirow{4}{*}{Empty string} 
& Baseline & 85.6 & 84.5 & 68.1 & 68.2 & 83.5 & 85.9 & 86.4 & 63.0 \\
& +COPRO & 87.2 & 85.9 & 67.1 & 67.0 & 83.3 & 85.4 & 85.5 & 60.8 \\
& +MiPROv2 & 85.0 & 83.8 & \textbf{69.2} & \textbf{69.0} & \textbf{86.5} & \textbf{88.4} & \textbf{87.7} & 62.2 \\
& +SIMBA & \textbf{87.8} & \textbf{86.8} & 66.7 & 66.4 & 84.1 & 86.1 & 86.7 & \textbf{63.1} \\
\midrule
\multirow{4}{*}{\makecell[l]{Check whether the table data supports the claim.}} 
& Baseline & 89.4 & 88.8 & 67.1 & 67.2 & 84.0 & 86.1 & 86.3 & 62.3 \\
& +COPRO & \textbf{90.0} & \textbf{89.5} & 66.4 & 66.4 & 84.4 & 86.6 & 85.4 & 62.2 \\
& +MiPROv2 & 88.3 & 87.5 & 66.4 & 66.2 & 84.2 & 86.4 & \textbf{86.9} & 63.2 \\
& +SIMBA & 89.4 & 88.5 & \textbf{68.5} & \textbf{68.2} & \textbf{86.2} & \textbf{87.6} & 86.6 & \textbf{63.8} \\
\midrule
\multirow{4}{*}{\parbox{9.5cm}{Examine the claim provided below and systematically compare each component of the statement against the corresponding values, entries, and relationships present in the table. Determine whether the claim is supported by, contradicts, or cannot be verified from the available tabular information.}} 
& Baseline & 83.9 & 83.0 & \textbf{68.8} & \textbf{68.8} & 85.9 & 87.9 & \textbf{87.0} & 62.9 \\
& +COPRO & \textbf{87.8} & \textbf{86.9} & \textbf{68.8} & \textbf{68.8} & 85.5 & 87.6 & 86.9 & \textbf{63.4} \\
& +MiPROv2 & 85.6 & 84.8 & 66.2 & 66.2 & \textbf{87.0} & \textbf{88.7} & \textbf{87.0} & 63.1 \\
& +SIMBA & \textbf{87.8} & 86.8 & 65.7 & 65.8 & 85.9 & 87.5 & 85.8 & 62.7 \\
\bottomrule
\end{tabular}
}
\caption{Ablation study on seed instructions for Qwen3-32B with CoT. \textbf{Bold} indicates best performance per seed instruction and dataset. \xt{TODO: improve size and readability; the instruction texts look too small}}
\label{tab:qwen3_cot_seed_instruction_ablation}
\end{table*}

\subsection{Optimizing ReAct with TART tools}
\begin{table*}[ht!]
\centering
\small
\scalebox{1.0}{
\begin{tabular}{llcccccccc}
\toprule
& & \multicolumn{2}{c}{\textbf{PubHealth}} & \multicolumn{2}{c}{\textbf{SciTab}} & \multicolumn{2}{c}{\textbf{TabFact}} & \multicolumn{2}{c}{\textbf{MMSci}} \\
\cmidrule(lr){3-4} \cmidrule(lr){5-6} \cmidrule(lr){7-8} \cmidrule(lr){9-10}
\textbf{Training Data} & \textbf{Optimizer} & Acc & F1 & Acc & F1 & Acc & F1 & Acc & F1 \\
\midrule
\multirow{4}{*}{Hybrid} 
& Baseline & 88.3 & 87.6 & 66.4 & 66.4 & 84.5 & 86.6 & 86.5 & 61.6 \\
& +COPRO & 87.2 & 86.1 & 67.4 & 67.3 & 85.5 & 87.6 & 86.7 & 61.5 \\
& +MiPROv2 & 87.2 & 86.5 & \textbf{68.8} & \textbf{68.6}& \textbf{86.9} & \textbf{88.5} & \textbf{87.7} & \textbf{65.4} \\
& +SIMBA & \textbf{90.0} & \textbf{89.6} & \textbf{68.8} & \textbf{68.6} & 85.2 & 87.1 & 87.0 & 64.2 \\
\midrule
\multirow{4}{*}{SciTab} 
& +COPRO & 87.2 & 86.5 & \textbf{69.7} & \textbf{69.8} & 83.2 & 85.4 & 85.5 & 62.4 \\
& +MiPROv2 & \textbf{88.9} & \textbf{88.0} & 68.3 & 68.2 & \textbf{86.5} & \textbf{88.3} & 87.4 & 62.1 \\
& +SIMBA & 87.8 & 87.3 & 64.1 & 63.7 & 84.8 & 86.5 & \textbf{88.3} & \textbf{67.4} \\
\midrule
\multirow{3}{*}{TabFact}
& +COPRO & 88.3 & 88.8 & 67.4 & 69.2 & 85.4 & 87.0 & 86.4 & 87.7 \\
& +MiPROv2 & 87.2 & 88.3 & \textbf{70.2} & \textbf{71.3} & 86.1 & 88.1 & 87.0 & 88.9 \\
& +SIMBA & \textbf{90.0} & \textbf{89.3} & 64.3 & 67.5 & \textbf{87.0} & \textbf{88.2} & \textbf{88.2} & \textbf{89.0} \\
\midrule
\multirow{3}{*}{PubHealth}
& +COPRO & 87.8 & 87.3 & \textbf{68.5} & \textbf{68.4} & \textbf{86.6} & \textbf{88.1} & \textbf{87.8} & \textbf{65.8} \\
& +MiPROv2 & 87.2 & 86.4 & 66.0 & 66.0 & 85.6 & 87.6 & 87.4 & \textbf{65.8} \\
& +SIMBA & \textbf{88.9} & \textbf{88.3} & 67.4 & 67.3 & 85.4 & 87.1 & 86.6 & 62.8 \\
\bottomrule
\end{tabular}
}
\caption{Ablation study on training data for Qwen3-32B with CoT. \textbf{Bold} indicates best performance per training data configuration and dataset.}
\label{tab:qwen3_cot_training_data_ablation}
\end{table*}

In addition to SQL tool, we investigate ReAct agents equipped with single or multiple tools that are most commonly used in TART framework \cite{lu-etal-2025-tart}.
We optimize the instructions in ReAct agents with Qwen3-32B as the backbone model using MiPROv2 and SIMBA, since these optimizers are more effective for improving ReAct performance based on \Cref{tab:qwen3_combined_results}. 
As shown in \Cref{tab:qwen3_react_tool_functions_ablation}, increasing the number of TART tools does not necessarily lead to better optimized performance for ReAct agents. 
In most experiment settings, F1 scores on the test sets after optimization follow a U-shaped trend as the number of tool functions is increased from 3 to 10.
ReAct with a single tool such as \texttt{get\_column\_cell\_value} can achieve stronger performance before and after instruction optimization than using top 10 TART tools.
The \texttt{get\_row\_by\_name} tool is particularly effective on PubHealthTab and MMSci, while \texttt{get\_row\_index\_by\_value} demonstrates strong performance on SciTab. 
Despite showing effectiveness in some evaluation settings, ReAct with TART tools or their combinations does not outperform the counterpart with SQL tool after SIMBA optimization. 


\subsection{Training Data Configuration}

\begin{table*}[ht!]
\centering
\small
\scalebox{1.0}{
\begin{tabular}{llcccccccc}
\toprule
& & \multicolumn{2}{c}{\textbf{PubHealth}} & \multicolumn{2}{c}{\textbf{SciTab}} & \multicolumn{2}{c}{\textbf{TabFact}} & \multicolumn{2}{c}{\textbf{MMSci}} \\
\cmidrule(lr){3-4} \cmidrule(lr){5-6} \cmidrule(lr){7-8} \cmidrule(lr){9-10}
\textbf{Tool Functions} & \textbf{Optimizer} & Acc & F1 & Acc & F1 & Acc & F1 & Acc & F1 \\
\midrule
\multirow{3}{*}{execute\_sql} 
& Baseline & 87.8 & 87.4 & 61.5 & 60.1 & 82.8 & 83.3 & \textbf{87.5} & 62.6\\
& +MiPROv2 & 87.8 & 87.2 & 61.5 & 60.9 & 84.2 & 85.2 & 86.2 & 63.0 \\
& +SIMBA & \textbf{90.6} & \textbf{90.0} & \textbf{66.2} & \textbf{65.9} & \textbf{86.1} & \textbf{87.0} & 85.9 & \textbf{65.0} \\
\midrule
\multirow{3}{*}{Top 10 TART tools} 
& Baseline & 85.0 & 84.3 & 59.4 & 58.1 & 83.4 & 84.8 & 84.7 & 61.8 \\
& +MiPROv2 & \textbf{87.8} & \textbf{87.1} & 60.1 & 59.3 & \textbf{86.0} & \textbf{87.4} & \textbf{87.8} & \textbf{62.4} \\
& +SIMBA & 84.4 & 83.9 & \textbf{62.7} & \textbf{62.7} & 81.6 & 84.4 & 81.6 & 57.8 \\
\midrule
\multirow{3}{*}{Top 3 TART tools}
& Baseline & 85.0 & 84.1 & 58.0 & 56.6 & 83.4 & 85.0 & 83.2 & 59.1 \\
& +MiPROv2 & \textbf{86.7} & \textbf{85.9} & \textbf{61.3} & \textbf{60.2} & \textbf{83.5} & \textbf{85.9} & \textbf{86.6} & \textbf{62.5} \\
& +SIMBA & 84.4 & 83.9 & 58.7 & 58.5 & 82.1 & 84.2 & 82.8 & 58.6 \\
\midrule
\multirow{3}{*}{Top 5 TART tools}
& Baseline & 84.4 & 83.6 & \textbf{59.7} & 58.0 & \textbf{84.1} & 85.6 & 83.8 & 59.7 \\
& +MiPROv2 & \textbf{85.0} & \textbf{84.2} & 58.3 & 57.2 & 83.9 & \textbf{85.8} & \textbf{86.3} & \textbf{62.6} \\
& +SIMBA & 82.8 & 81.6 & 58.7 & \textbf{58.1} & 82.0 & 84.0 & 82.2 & 58.1 \\
\midrule
\multirow{3}{*}{equal\_to}
& Baseline & \textbf{86.7} & \textbf{85.9} & 59.9 & 59.6 & 83.6 & 84.5 & 86.1 & 62.8 \\
& +MiPROv2 & 86.1 & 85.0 & 59.4 & 58.5 & \textbf{84.3} & \textbf{85.3} & \textbf{87.3} & \textbf{63.4} \\
& +SIMBA & 85.0 & 84.7 & \textbf{66.0} & \textbf{65.8} & 83.5 & 84.8 & 85.8 & 61.4 \\
\midrule
\multirow{3}{*}{get\_column\_by\_name}
& Baseline & 87.2 & 86.6 & 61.8 & 60.7 & \textbf{85.2} & \textbf{86.6} & \textbf{85.8} & 60.8 \\
& +MiPROv2 & \textbf{87.8} & \textbf{87.2} & 59.2 & 58.6 & 81.6 & 84.4 & 84.8 & \textbf{61.7} \\
& +SIMBA & 83.9 & 82.5 & \textbf{63.9} & \textbf{64.1} & 83.4 & 85.7 & 83.7 & 60.2 \\
\midrule
\multirow{3}{*}{get\_column\_cell\_value}
& Baseline & 87.8 & 86.6 & 59.9 & 58.5 & 83.3 & 85.1 & 84.4 & \textbf{62.1} \\
& +MiPROv2 & \textbf{88.3} & \textbf{87.8} & 60.4 & 59.7 & \textbf{84.9} & \textbf{86.2} & \textbf{86.8} & 61.8 \\
& +SIMBA & \textbf{88.3} & 87.6 & \textbf{65.3} & \textbf{65.1} & 81.7 & 84.0 & 81.4 & 59.4 \\
\midrule
\multirow{3}{*}{get\_row\_by\_name}
& Baseline & 88.3 & 87.7 & 57.8 & 55.6 & 85.2 & 86.6 & 87.0 & 63.6 \\
& +MiPROv2 & \textbf{88.9} & \textbf{88.2} & 60.1 & 58.5 & \textbf{85.4} & 86.7 & \textbf{88.1} & 63.0 \\
& +SIMBA & 87.2 & 86.3 & \textbf{61.1} & \textbf{60.7} & 85.0 & \textbf{86.8} & 87.5 & \textbf{66.6} \\
\midrule
\multirow{3}{*}{get\_row\_index\_by\_value}
& Baseline & \textbf{87.8} & \textbf{87.0} & 62.2 & 61.6 & \textbf{84.6} & \textbf{86.5} & 86.5 & \textbf{63.1} \\
& +MiPROv2 & 87.2 & 86.5 & 64.1 & 63.8 & 82.6 & 84.4 & \textbf{86.8} & 61.6 \\
& +SIMBA & 87.2 & 86.7 & \textbf{66.0} & \textbf{66.1} & 82.4 & 85.6 & 86.5 & 61.2 \\
\bottomrule
\end{tabular}
}
\caption{Ablation study on tool functions used in TART for Qwen3-32B with ReAct. \textbf{Bold} indicates best performance per tool function and dataset.}
\label{tab:qwen3_react_tool_functions_ablation}
\end{table*}

We investigate the impact of training data on instruction optimization by conducting the optimization with three training configurations.
Instead of using hybrid train data, we randomly sample 100 instances from each of the SciTab, TabFact and PubHealthTab training sets to construct single-source training sets.
As shown in \Cref{tab:qwen3_cot_training_data_ablation}, SIMBA benefits more from using a hybrid train set, while MiPROv2 is less sensitive to the distribution of train data.
This indicates hybrid data can help SIMBA learn more generalized rules across different table-based fact checking tasks.
In contrast, MiPROv2 follows human-written heuristic rules to propose new candidate instructions and mainly relies on the train data to understand the task format.
Also, we find optimized instructions learned on TabFact data can generalize surprisingly well on SciTab and MMSci, probably due to cleaner data and binary label distribution in TabFact. 
%

\xt{TODO: summarize the findings}

\xt{No need to cut this section; you can just explain what you did and what you observed; briefly provide some hypothesis about why the performance when using TabFact as train data is much higher on MMSci and PubHealthTab}

\begin{table*}[]
\centering
\resizebox{0.95\linewidth}{!}{
\begin{tabular}{l}
\toprule
\multicolumn{1}{c}{\textbf{Ordinal Terms}} \\
\midrule
\begin{tabular}[c]{@{}l@{}}`first', `second', `third', `fourth', `fifth', `sixth', `seventh', `eighth', `ninth', `tenth', `next', `last', \\`previous', `following', `initial', `final', `primary', `secondary', `subsequent',  `preceding', `beginning', \\ `middle', `end', `earlier', `later', `prior', `posterior', `successive', `consecutive', `sequential'\end{tabular}\\
\bottomrule
\end{tabular}
}
\caption{List of ordinal terms used in the experiments in \Cref{sec:analysis_baseline_cot}}
\label{tab:ordinal_terms}
\end{table*}

\begin{figure}[htbp]
\centering
\begin{tcolorbox}[colback=gray!10!white,colframe=black!50!black,title=System prompt (Two classes),fonttitle=\bfseries, halign title=flush center, width=0.5\textwidth]
{You verify claims against tables step-by-step using SQL, Python, or a direct answer.\\
\\
RULES:\\
1. Each step: pick action\_type = SQL | Python | Answer.\\
2. SQL/Python → fill "code" with executable code. NEVER put a label in "code".\\
3. Answer → fill "label" with exactly "supported" or "refuted".\\
4. Table name is DF. Wrap column names with spaces in backticks: `Col Name`.\\
5. An executor runs your code and returns an intermediate table.\\
6. Keep code short (<300 chars). Operate on the most recent intermediate table.\\
\\
IMPORTANT: "code" = executable SQL/Python ONLY. "label" = final verdict ONLY. Never mix them.\\
}
\end{tcolorbox}
\caption{System prompt used for two-class table-based fact checking tasks.}
\label{system_prompt_two_class}
\end{figure}
\begin{figure}[htbp]
\centering
\begin{tcolorbox}[colback=gray!10!white,colframe=black!50!black,title=System prompt (Three classes),fonttitle=\bfseries, halign title=flush center, width=0.5\textwidth]
{You verify claims against tables step-by-step using SQL, Python, or a direct answer.\\

RULES:\\
1. Each step: pick action\_type = SQL | Python | Answer.\\
2. SQL/Python → fill "code" with executable code. NEVER put a label in "code".\\
3. Answer → fill "label" with exactly "supported", "refuted", or "not verifiable".\\
4. Table name is DF. Wrap column names with spaces in backticks: `Col Name`.\\
5. An executor runs your code and returns an intermediate table.\\
6. Keep code short (<300 chars). Operate on the most recent intermediate table.\\
\\
Labels:\\
- "supported": table confirms the claim.\\
- "refuted": table contradicts the claim.\\
- "not verifiable": table lacks relevant info (only after examining data).\\
\\
IMPORTANT: "code" = executable SQL/Python ONLY. "label" = final verdict ONLY. Never mix them.\\
}
\end{tcolorbox}
\caption{System prompt used for three-class table-based fact checking tasks.}
\label{system_prompt_three_class}
\end{figure}
\begin{figure}[htbp]
\centering
\begin{tcolorbox}[colback=gray!10!white,colframe=black!50!black,title=Task prompt (Two classes),fonttitle=\bfseries, halign title=flush center, width=0.5\textwidth]
{The database table DF is shown as follows:\\
Caption:\{table caption\}\\
\{table\}\\
\\
Answer the following question based on the data above: ``\{claim\}'' Is the above claim supported or refuted by the provided table? Generate SQL or Python code step-by-step given the question and table to answer the question correctly. For each step, generate SQL code to process the query or Python code to reformat the data. Output the code braced by ``\textasciigrave\textasciigrave\textasciigrave'' and an external executor will process the code generated and feed an intermediate table back to you. Answer the question directly if confident.
}
\end{tcolorbox}
\caption{Task prompt template used for two-class table-based fact checking tasks.}
\label{task_prompt_two_class}
\end{figure}
\begin{figure}[htbp]
\centering
\begin{tcolorbox}[colback=gray!10!white,colframe=black!50!black,title=Task prompt (Three classes),fonttitle=\bfseries, halign title=flush center, width=0.5\textwidth]
{The database table DF is shown as follows:\\
Caption:\{table caption\}\\
\{table\}\\
\\
Answer the following question based on the data above: ``\{claim\}'' Is the above claim supported, refuted or not verifiable based on the provided table? Generate SQL or Python code step-by-step given the question and table to answer the question correctly. For each step, you can choose to generate SQL code to process the query or Python code to reformat the data, or answer the question directly with explanations. Output the code braced by \"```\" and an external executor will process the code generated and feed an intermediate table back to you. Answer the question directly if confident.\\
Answer with exactly one of these labels: supports, refutes, or not enough info.
}
\end{tcolorbox}
\caption{Task prompt template used for three-class table-based fact checking tasks.}
\label{task_prompt_three_class}
\end{figure}

\begin{table*}[]
\centering
\resizebox{0.95\linewidth}{!}{
\begin{tabular}{l}
\toprule
\multicolumn{1}{c}{\textbf{Comparative Terms}} \\
\midrule
\begin{tabular}[c]{@{}l@{}}'lower', 'higher', 'greater', 'smaller', 'larger', 'lesser', 'equal', 'unequal', 'bigger', 'shorter', \\ 'longer', 'deeper', 'stronger', 'weaker', 'faster', 'slower', 'earlier', 'later', 'better', 'worse', \\ 'superior', 'inferior', 'maximum', 'minimum', 'greater than', 'less than', 'equal to', 'more than', \\ 'fewer than', 'at least', 'at most', 'not more than', 'not less than'\end{tabular}\\
\bottomrule
\end{tabular}
}
\caption{List of comparative terms used in the experiments in \Cref{sec:analysis_baseline_cot}}
\label{tab:comp_terms}
\end{table*}



\xt{Qualitative analysis on tool calling traces?}
\xt{Tables with tool use statistics}


\begin{table*}[]
\resizebox{\linewidth}{!}{
\begin{tabular}{lp{\textwidth}}
\toprule
\textbf{Optimizer} & \textbf{Instruction} \\
\midrule
Baseline & Verify the given claim against the provided table data. \\
\midrule
COPRO & Verify the given claim against the provided table data. \\
\midrule
MiPROv2 & Analyze the claim and the associated table data to determine whether the claim is supported, refuted, or neutral based on the information provided in the table. Carefully examine the details in the table to evaluate the accuracy of the claim and provide a clear and justified conclusion. \\
\midrule
SIMBA & Verify the given claim against the provided table data. 
\\
\bottomrule
\end{tabular}
}
\caption{Baseline and optimized instructions for the Qwen3-8B model with direct prompting using various optimizers.}
\label{tab:inst_qwen8b_direct}
\end{table*}

\begin{table*}[]
\small
\resizebox{\linewidth}{!}{
\begin{tabular}{lp{0.865\textwidth}}
\toprule
\textbf{Optimizer} & \textbf{Instruction} \\
\midrule
Baseline & Verify the given claim against the provided table data. \\
\midrule
COPRO & Analyze the table data and the claim thoroughly. Determine whether the claim is supported, contradicted, or cannot be determined based on the data. Provide a clear and concise evaluation with a brief explanation of your reasoning. \\
\midrule
MiPROv2 & Analyze the claim, table data, and table caption to determine whether the claim is supported, refuted, or if there is not enough information to verify it. Provide a step-by-step reasoning process explaining how you arrived at your conclusion, and clearly state your final answer as one of the following: 'supports', 'refutes', or 'not enough info'. \\
\midrule
SIMBA & Verify the given claim against the provided table data.\textbackslash{}n\textbackslash{}nIf the module receives a claim comparing the price values of different energy sources, it should focus on the actual price numbers provided in the table rather than interpreting the claim as referring to the count of prices. Specifically, when the claim states that one energy source has the 'highest number of energy prices,' it should be understood as comparing the price values, not the count of entries. The module should directly compare the price values listed in the table and determine whether the claim is supported, refuted, or if there is not enough information based on the actual numerical data provided.\textbackslash{}n\textbackslash{}nIf the module receives a claim that refers to a specific difference (e.g., 'lower by 1.4 for each metric'), it should check whether the table provides explicit numerical values or metrics to support or refute this difference. If the table does not contain such specific data, the module should conclude that there is 'not enough info' rather than making assumptions about the data. Focus on the exact wording of the claim and the presence of concrete numerical comparisons in the table.\textbackslash{}n\textbackslash{}nIf the module receives a claim comparing two models (e.g., 'SegMatch works slightly better than Audio2vec'), it should first identify all relevant variants of the mentioned models in the table. For each metric in the table, compare the values of the models directly. If the claim refers to a specific difference (e.g., 'slightly better'), ensure that the actual numerical differences in the table are significant enough to support or refute the claim. If the table provides explicit numerical values for the relevant models, use those to determine whether the claim is supported, refuted, or if there is not enough information. Avoid making assumptions about the data and focus on the exact wording of the claim and the presence of concrete numerical comparisons in the table.\\
\bottomrule
\end{tabular}
}
\caption{Baseline and optimized instructions for the Qwen3-8B model with CoT method using various optimizers.}
\label{tab:inst_qwen8b_cot}
\end{table*}

\begin{table*}[]
\small
\resizebox{\linewidth}{!}{
\begin{tabular}{lp{0.865\textwidth}}
\toprule
\textbf{Optimizer} & \textbf{Instruction} \\
\midrule
Baseline & Verify the given claim against the provided table data. \\
\midrule
COPRO & You are an Agent tasked with evaluating a claim against a given table. You will be provided with the `claim`, `table`, and `caption` as input. Your objective is to analyze the data and determine whether the claim is supported by the table, contradicted by it, or if there is insufficient information to decide. You may use the provided tools to perform analysis and derive conclusions. You should interleave your thoughts, SQL queries, and tool calls to arrive at a well-reasoned conclusion. \\
\midrule
MiPROv2 & You are an Agent tasked with verifying a claim against a table. You are provided with the `claim`, `table`, and `caption` as input, along with your past trajectory. Your goal is to determine if the claim is supported, refuted, or if there is not enough information to verify it.\textbackslash{}n\textbackslash{}nTo accomplish this, you will:\textbackslash{}n1. Carefully analyse the claim and the table data.\textbackslash{}n2. Use your reasoning to plan your next step, which could involve:\textbackslash{}n   a) Executing a SQL query to extract relevant data from the table using the `execute\_sql` tool. This tool takes the table data, table name, and SQL query as arguments.\textbackslash{}n   b) Concluding the task by using the `finish` tool when you have sufficient information to evaluate the claim.\textbackslash{}n3. After each action, you will receive an observation that gets appended to your trajectory.\textbackslash{}n4. Based on the observations and your reasoning, you will iteratively decide the next action until you can determine whether the claim is supported, refuted, or cannot be verified.\textbackslash{}n\textbackslash{}nWhen writing `next\_thought`, you must provide a clear explanation of your reasoning and plan for the next step. When selecting `next\_tool\_name`, make sure to choose between `execute\_sql` or `finish`. When specifying `next\_tool\_args`, ensure the arguments are provided in JSON format.\textbackslash{}n\textbackslash{}nYour final answer should be one of:\textbackslash{}n- \textbackslash{}"supports\textbackslash{}" if the claim is supported by the table data.\textbackslash{}n- \textbackslash{}"refutes\textbackslash{}" if the claim is refuted by the table data.\textbackslash{}n- \textbackslash{}"not enough info\textbackslash{}" if the table data does not provide sufficient information to evaluate the claim.\textbackslash{}n\textbackslash{}nKeep your reasoning process clear and concise, and make sure to justify your actions based on the information available in the table. \\
\midrule
SIMBA & Verify the given claim against the provided table data.\textbackslash{}n\textbackslash{}nYou are an Agent. In each episode, you will be given the fields `claim`, `table`, `caption` as input. And you can see your past trajectory so far.\textbackslash{}nYour goal is to use one or more of the supplied tools to collect any necessary information for producing `answer`.\textbackslash{}n\textbackslash{}nTo do this, you will interleave next\_thought, next\_tool\_name, and next\_tool\_args in each turn, and also when finishing the task.\textbackslash{}nAfter each tool call, you receive a resulting observation, which gets appended to your trajectory.\textbackslash{}n\textbackslash{}nWhen writing next\_thought, you may reason about the current situation and plan for future steps.\textbackslash{}nWhen selecting the next\_tool\_name and its next\_tool\_args, the tool must be one of:\textbackslash{}n\textbackslash{}n(1) execute\_sql, whose description is \textless{}desc\textgreater      Execute a SQL query on a table provided as list of lists format.            Args:          table\_data: List of lists where first row contains headers, subsequent rows contain data          table\_name: Name of the table dataframe for executing SQL query          sql\_query: SQL query string to execute            Returns:          Formatted string of the transformed table or error message      \textless{}/desc\textgreater{}. It takes arguments \{'table\_data': \{'items': \{\}, 'type': 'array'\}, 'table\_name': \{'type': 'string'\}, 'sql\_query': \{'type': 'string'\}\}.\textbackslash{}n(2) finish, whose description is \textless{}desc\textgreater{}Marks the task as complete. That is, signals that all information for producing the outputs, i.e. `answer`, are now available to be extracted.\textless{}/desc\textgreater{}. It takes arguments \{\}.\textbackslash{}nWhen providing `next\_tool\_args`, the value inside the field must be in JSON format\textbackslash{}n\textbackslash{}nIf the module receives a claim that involves comparing systems based on their training methodology (e.g., reinforcement learning vs. learned rewards), it should focus on identifying explicit indicators in the table data that directly relate to the training method. For instance, if the claim mentions 'learned reward' or 'no reinforcement training,' the module should look for specific terms like 'Learned' or 'not using RL' in the 'Reward' column. Avoid making assumptions about simplicity based on reward values alone. Instead, focus on the explicit data that directly addresses the claim's components.\textbackslash{}n\textbackslash{}nIf the module receives a claim that involves comparing systems based on their training methodology (e.g., reinforcement learning vs. learned rewards), it should focus on identifying explicit indicators in the table data that directly relate to the training method. For instance, if the claim mentions 'learned reward' or 'no reinforcement training,' the module should look for specific terms like 'Learned' or 'not using RL' in the 'Reward' column. Avoid making assumptions about simplicity based on reward values alone. Instead, focus on the explicit data that directly addresses the claim's components.\\
\bottomrule
\end{tabular}
}
\caption{Baseline and optimized instructions for the Qwen3-8B model with ReAct method using various optimizers.}
\label{tab:inst_qwen8b_react}
\end{table*}

\begin{table*}[]
\small
\resizebox{\linewidth}{!}{
\begin{tabular}{lp{0.865\textwidth}}
\toprule
\textbf{Optimizer} & \textbf{Instruction} \\
\midrule
Baseline & Verify the given claim against the provided table data. \\
\midrule
COPRO & You are an intelligent agent tasked with verifying a claim against a provided table. You will receive the following inputs: `claim` (a statement to be verified), `table` (a dataset containing rows and columns), and `caption` (a description of the table). Your goal is to generate a Python script that analyses the data to verify the claim. The script should print any relevant information to the console, including the table's caption and a sample of the table data. Once all necessary information is collected and the claim is verified, mark the task as completed with `finished=True` and provide a clear final answer. \\
\midrule
MiPROv2 & Verify the given claim against the provided table data. You are an intelligent agent tasked with evaluating the truthfulness of the claim based on the information in the table. For each episode, you will receive the fields `claim`, `table`, and `caption` as input.\textbackslash{}n\textbackslash{}nYour goal is to generate executable Python code that processes the table data to determine whether the claim is supported, refuted, or not enough information is available. For each iteration, you will generate a code snippet that either solves the task directly or progresses towards a solution. Ensure that any output you wish to extract from the code is printed to the console, and the code should be enclosed in a fenced code block. When all necessary information for determining the truth of the claim is available, mark `finished=True` to indicate the completion of the process.\textbackslash{}n\textbackslash{}nYou have access to the Python Standard Library and can use any functions provided to analyze the table data. Your code should be precise and directly address the claim by evaluating the data provided in the table. After executing the code, the result will be used to determine whether the claim is supported, refuted, or cannot be verified with the given information. \\
\midrule
SIMBA & Verify the given claim against the provided table data.\textbackslash{}n\textbackslash{}nYou are an intelligent agent. For each episode, you will receive the fields `claim`, `table`, `caption` as input.\textbackslash{}nYour goal is to generate executable Python code that collects any necessary information for producing `answer`.\textbackslash{}nFor each iteration, you will generate a code snippet that either solves the task or progresses towards the solution.\textbackslash{}nEnsure any output you wish to extract from the code is printed to the console. The code should be enclosed in a fenced code block.\textbackslash{}nWhen all information for producing the outputs (`answer`) are available to be extracted, mark `finished=True` besides the final Python code.\textbackslash{}nYou have access to the Python Standard Library and the following functions:\textbackslash{}n\textbackslash{}nIf the module receives a claim that involves specific events or scenarios not directly mentioned in the table data, it should carefully analyze the table's content to determine if there is any indirect or related information that could support or refute the claim. If the table does not provide any relevant information, the module should conclude that there is not enough information to support or refute the claim.\textbackslash{}n\textbackslash{}nIf the module receives a claim that involves specific features or embeddings, it should directly compare the relevant metrics for those features rather than averaging them. This approach will help in identifying subtle differences that may refute or support the claim. Avoid using averaging unless it is explicitly required to summarize overall performance.\textbackslash{}n\textbackslash{}nIf the module receives a claim and table data, it should first parse the table data into a structured format (e.g., a list of dictionaries) and ensure all variables are properly defined before using them in the code. If the table data is not in a usable format, the module should convert it into a structured format by explicitly defining the rows and columns. Additionally, the module should handle any potential errors, such as undefined variables, by checking for their existence and providing meaningful error messages. This will prevent incorrect conclusions based on faulty data parsing.\textbackslash{}n\textbackslash{}nIf the module receives a claim that involves specific features or embeddings, it should directly compare the relevant metrics for those features rather than averaging them. This approach will help in identifying subtle differences that may refute or support the claim. Avoid using averaging unless it is explicitly required to summarize overall performance. Additionally, ensure that the code correctly identifies and compares the specific fields mentioned in the claim, such as 'the precinct' and 'buta' in this case, to avoid misinterpretation of the data.
\\
\bottomrule
\end{tabular}
}
\caption{Baseline and optimized instructions for the Qwen3-8B model with CodeAct method using various optimizers.}
\label{tab:inst_qwen8b_codeact}
\end{table*}

\begin{table*}[]
\small
\resizebox{\linewidth}{!}{
\begin{tabular}{lp{0.865\textwidth}}
\toprule
\textbf{Optimizer} & \textbf{Instruction} \\
\midrule
Baseline & Verify the given claim against the provided table data. \\
\midrule
COPRO & Given a table of data, evaluate the truthfulness of a specific claim by thoroughly cross-referencing it with the information provided in the table. Explain the reasoning behind your conclusion clearly. \\
\midrule
MiPROv2 & Verify the given claim against the provided table data. \\
\midrule
SIMBA & Verify the given claim against the provided table data.\textbackslash{}n\textbackslash{}nIf the module receives a claim that specifies a particular treatment as the 'first' or 'second' alternative for a given condition, it should carefully cross-check the table data to ensure the claim's order of alternatives matches the table. Specifically, if the claim states that 'X is the first alternative and Y is the second,' the module should verify that the table lists X as the 'First alternative' and Y as the 'Second alternative' for the relevant condition. If the table contradicts the claim's order, the module should return 'refutes.'\textbackslash{}n\textbackslash{}nIf the module receives a claim that specifies a particular condition (e.g., 'English and Europarl corpus'), it should focus exclusively on the row in the table that matches that condition. It should then compare the precision values of DocSub and DF in that row. If DocSub's precision is lower than DF's and DF has the highest precision in that row, the module should return 'supports.' Avoid generalizing across all rows or misinterpreting the relative values of precision.\textbackslash{}n\textbackslash{}nIf the module receives a claim that discusses a system's training method (e.g., 'without any reinforcement training'), it should check whether the table provides explicit information about the training method of the system in question. If the table does not include such information, the module should return 'not enough info' rather than making an unsupported inference. Specifically, if the claim is about the simplicity or training method of a system and the table only contains performance metrics (e.g., ROUGE scores), the module should avoid drawing conclusions about the training method and instead return 'not enough info.'\textbackslash{}n\textbackslash{}nIf the module receives a claim that involves combining the points of players from the same club, it should sum the points for each club and compare the totals. Specifically, if the claim states that 'kãlner haie gets to be the leading in points,' the module should calculate the total points for kãlner haie and compare it with the total points for other clubs. If kãlner haie's total points are not the highest, the module should return 'refutes.' Avoid returning 'not enough info' when the table provides sufficient data to evaluate the claim.
\\
\bottomrule
\end{tabular}
}
\caption{Baseline and optimized instructions for the Qwen3-32B model with direct prompting using various optimizers.}
\label{tab:inst_qwen32b_direct}
\end{table*}

\begin{table*}[]
\small
\resizebox{\linewidth}{!}{
\begin{tabular}{lp{0.865\textwidth}}
\toprule
\textbf{Optimizer} & \textbf{Instruction} \\
\midrule
Baseline & Verify the given claim against the provided table data. \\
\midrule
COPRO & Given a claim and a table of data, assess whether the claim is *completely supported*, *partially supported*, *contradicted*, or *inconclusive* based on the information in the table. Provide a concise yet thorough explanation for your evaluation. Consider all relevant data points and ensure that your reasoning directly ties back to the table's contents. \\
\midrule
MiPROv2 & Analyze the given claim by thoroughly examining the provided table and its caption. Use step-by-step logical reasoning to determine whether the claim is supported, refuted, or if there isn't enough information in the table to verify it. Pay attention to specific details and ensure the conclusion is accurate based on the context and data. \\
\midrule
SIMBA & Verify the given claim against the provided table data.\textbackslash{}n\textbackslash{}nIf the claim refers to the effectiveness or performance of a method in a specific stage, and the table includes evaluation metrics (e.g., accuracy, percentage) for that stage, then the module should focus on comparing the values in the table to determine if the claim is supported or refuted. Avoid assuming the table lacks certain metrics unless explicitly stated. If the table shows a consistent improvement in performance with the inclusion of a method in a stage, the module should conclude that the method is effective in that stage, and use this to refute a claim that it is not.
\\
\bottomrule
\end{tabular}
}
\caption{Baseline and optimized instructions for the Qwen3-32B model with CoT method using various optimizers.}
\label{tab:inst_qwen32b_cot}
\end{table*}

\begin{table*}[]
\small
\resizebox{\linewidth}{!}{
\begin{tabular}{lp{0.865\textwidth}}
\toprule
\textbf{Optimizer} & \textbf{Instruction} \\
\midrule
Baseline & Verify the given claim against the provided table data. \\
\midrule
COPRO & The proposed instruction is as follows:\textbackslash{}n\textbackslash{}n**Instruction:** You are a reasoning agent responsible for determining whether a provided `claim` is supported, inaccurate, or indeterminate based on the data in the provided `table` and its contextual `caption`. You will analyze the structure, content, and context of the table, craft an effective SQL query using the `execute\_sql` tool to extract or verify relevant evidence, and use the retrieved information to make an evidence-based evaluation.\textbackslash{}n\textbackslash{}nFollow the structured approach:\textbackslash{}n1. Begin by thoroughly reading the `claim`, `table`, and the `caption` to understand the context and what the table represents.\textbackslash{}n2. Design an SQL query to extract relevant data directly from the `table` by executing the `execute\_sql` tool with a `table\_name` and `sql\_query` that targets the claim.\textbackslash{}n3. Examine the query result closely and evaluate it in light of the claim.\textbackslash{}n4. Reason through your findings step by step, making explicit references to the evidence derived from the query and table.\textbackslash{}n5. Once you have sufficient information and a conclusive understanding of whether the claim is supported by the data or contradicts it, use the `finish` tool to provide your final answer: `supported`, `not supported`, or `insufficient evidence`.\textbackslash{}n\textbackslash{}nYour answer should be data-driven and well-supported, with clear reasoning. \\
\midrule
MiPROv2 & Carefully verify the accuracy of the provided claim using the structured table data. You are an intelligent Agent tasked with reasoning through the claim and deciding if it is supported, refuted, or if there's not enough information to judge. Use the `execute\_sql` tool to query the table when necessary for deeper insights and the `finish` tool once you are confident in your verification. Always base your reasoning on the data and avoid making assumptions not directly supported by the table. \\
\midrule
SIMBA & Verify the given claim against the provided table data.\textbackslash{}n\textbackslash{}nYou are an Agent. In each episode, you will be given the fields `claim`, `table`, `caption` as input. And you can see your past trajectory so far.\textbackslash{}nYour goal is to use one or more of the supplied tools to collect any necessary information for producing `answer`.\textbackslash{}n\textbackslash{}nTo do this, you will interleave next\_thought, next\_tool\_name, and next\_tool\_args in each turn, and also when finishing the task.\textbackslash{}nAfter each tool call, you receive a resulting observation, which gets appended to your trajectory.\textbackslash{}n\textbackslash{}nWhen writing next\_thought, you may reason about the current situation and plan for future steps.\textbackslash{}nWhen selecting the next\_tool\_name and its next\_tool\_args, the tool must be one of:\textbackslash{}n\textbackslash{}n(1) execute\_sql, whose description is \textless{}desc\textgreater      Execute a SQL query on a table provided as list of lists format.            Args:          table\_data: List of lists where first row contains headers, subsequent rows contain data          table\_name: Name of the table dataframe for executing SQL query          sql\_query: SQL query string to execute            Returns:          Formatted string of the transformed table or error message      \textless{}/desc\textgreater{}. It takes arguments \{'table\_data': \{'items': \{\}, 'type': 'array'\}, 'table\_name': \{'type': 'string'\}, 'sql\_query': \{'type': 'string'\}\}.\textbackslash{}n(2) finish, whose description is \textless{}desc\textgreater{}Marks the task as complete. That is, signals that all information for producing the outputs, i.e. `answer`, are now available to be extracted.\textless{}/desc\textgreater{}. It takes arguments \{\}.\textbackslash{}nWhen providing `next\_tool\_args`, the value inside the field must be in JSON format\textbackslash{}n\textbackslash{}nIf the table data explicitly contains the information needed to verify the claim, the module should immediately call the `finish` tool without executing unnecessary SQL queries. Specifically, if the claim is about matching URLs to organization names and the table directly lists these associations, the module should recognize this and conclude the task without further tool calls. Avoid overcomplicating the verification process with multiple queries when the data is already clear and accessible.\textbackslash{}n\textbackslash{}nIf the table data explicitly contains the required information for verification, the module should avoid executing unnecessary SQL queries. Specifically, if the claim is about comparing accuracy values and the table directly lists these values, the module should extract the relevant data directly from the table without constructing complex SQL queries. Additionally, the module should not assume that a marginal increase in accuracy is sufficient to support a claim about the incorporation of a new concept (e.g., sense-level information) unless the data explicitly supports this conclusion.\textbackslash{}n\textbackslash{}nIf the claim refers to a specific metric (e.g., accuracy) and a specific setting (e.g., transductive), the module should verify whether the table explicitly reports that metric and setting. If the table only provides related metrics (e.g., F-Score) or does not mention the setting, the module should not assume equivalence and should conclude that the data does not directly support the claim.
\\
\bottomrule
\end{tabular}
}
\caption{Baseline and optimized instructions for the Qwen3-32B model with ReAct method using various optimizers.}
\label{tab:inst_qwen32b_react}
\end{table*}

\begin{table*}[]
\small
\resizebox{\linewidth}{!}{
\begin{tabular}{lp{0.865\textwidth}}
\toprule
\textbf{Optimizer} & \textbf{Instruction} \\
\midrule
Baseline & Verify the given claim against the provided table data. \\
\midrule
COPRO & You will act as a data verification assistant. Your task is to evaluate whether the provided `claim` is supported by the data in the `table` (a dictionary with key-value structure), which is accompanied by the `caption` (a brief explanation of the table's contents). Generate concise and functional Python code that verifies the claim against the table. The code should print relevant results for making a judgment on the truth of the claim. When the data needed to evaluate the claim is fully extracted or the task is complete, conclude by marking `finished=True` and output your findings in the answer field. \\
\midrule
MiPROv2 & Verify the accuracy of a given claim by generating Python code that interprets the provided table data, including checking for numerical trends, comparisons, and contextual details from the table caption.\textbackslash{}n\textbackslash{}nAs an intelligent agent, you will receive a claim, a table, and a caption as inputs. Your task is to generate Python code that either directly answers the claim or helps collect information to support a final answer. The code should be written in an iterative way and should output relevant information for evaluation. When the claim can be definitively answered, indicate this with `finished=True` alongside your code. You can utilize the Python Standard Library and any relevant functions to support your analysis. Make sure the output is printed clearly and in a structured format. \\
\midrule
SIMBA & Verify the given claim against the provided table data.\textbackslash{}n\textbackslash{}nYou are an intelligent agent. For each episode, you will receive the fields `claim`, `table`, `caption` as input.\textbackslash{}nYour goal is to generate executable Python code that collects any necessary information for producing `answer`.\textbackslash{}nFor each iteration, you will generate a code snippet that either solves the task or progresses towards the solution.\textbackslash{}nEnsure any output you wish to extract from the code is printed to the console. The code should be enclosed in a fenced code block.\textbackslash{}nWhen all information for producing the outputs (`answer`) are available to be extracted, mark `finished=True` besides the final Python code.\textbackslash{}nYou have access to the Python Standard Library and the following functions:\textbackslash{}n\textbackslash{}nIf the module receives a claim about the methodology of a classifier (e.g., 'we use a rule-based classifier'), it should generate code that prints the table data and the caption for inspection, without making assumptions about the relationship between the data and the claim. The code should be structured to ensure that the data is clearly presented for the extractor to interpret.\textbackslash{}n\textbackslash{}nIf the module receives a claim about a general trend (e.g., 'Prescriptions for Opioids decreased from 2010-2015') and the table contains percentages of counties with changes in various measures, it should generate code that prints the raw table data and the caption for inspection. Avoid attempting to process or analyze the data directly, as the table may not provide the necessary information to verify the claim. Ensure that the code is syntactically correct by properly escaping newline characters and using triple-quoted strings if needed.\textbackslash{}n\textbackslash{}nIf the module receives a claim about a chronological relationship between two events and the table contains the years of these events, it should generate code that correctly parses the years and checks the chronological order. Ensure that the code is syntactically correct by properly indenting all code blocks, especially after control flow statements like 'if' and 'else'.\textbackslash{}n\textbackslash{}nIf the module receives a claim about a general trend and the table contains percentages of counties with changes in various measures, it should generate code that prints the raw table data and the caption for inspection. Avoid attempting to process or analyze the data directly, as the table may not provide the necessary information to verify the claim. Ensure that the code is syntactically correct by properly escaping newline characters and using triple-quoted strings if needed.
\\
\bottomrule
\end{tabular}
}
\caption{Baseline and optimized instructions for the Qwen3-32B model with CodeAct method using various optimizers.}
\label{tab:inst_qwen32b_codeact}
\end{table*}

\begin{table*}[]
\small
\resizebox{\linewidth}{!}{
\begin{tabular}{lp{0.865\textwidth}}
\toprule
\textbf{Optimizer} & \textbf{Instruction} \\
\midrule
Baseline & Verify the given claim against the provided table data. \\
\midrule
COPRO & You are a precise and reliable fact-checking assistant. You will be provided with a claim and a table of data. Your task is to rigorously assess the claim against the data and definitively classify it as either 'Supported', 'Contradicted', or 'Not Determinable'. Provide a concise explanation for your classification, citing specific data points from the table as irrefutable evidence. Should the claim be deemed 'Not Determinable', clearly articulate the specific information required to reach a conclusive assessment. Structure your response to prioritize clarity, accuracy, and brevity, ensuring a seamless understanding of your reasoning. \\
\midrule
MiPROv2 & You are a fact-checking assistant. Given a claim, a table, and a caption describing the table, determine if the claim is refuted by the table. Return \textbackslash{}"refutes\textbackslash{}" if the claim contradicts information in the table, and \textbackslash{}"supports\textbackslash{}" otherwise. Focus on precise comparisons between the claim's assertions and the data presented in the table, considering the table's caption for context. \\
\midrule
SIMBA & Verify the given claim against the provided table data.\textbackslash{}n\textbackslash{}nWhen evaluating a claim involving numerical comparisons against table data, carefully extract the relevant numerical values from the table.  Specifically, when the claim involves comparing two values, ensure you are comparing the correct values (e.g., comparing 'conversion factor' for 'Hydrocodone' and 'Hydromorphone').  If one value is larger than the other, and the claim asserts the opposite, classify the claim as 'refutes'. Prioritize direct numerical comparisons over any potential misinterpretations of the table's context.\textbackslash{}n\textbackslash{}nWhen evaluating claims involving numerical comparisons against table data, prioritize direct numerical comparisons. Specifically, extract the relevant numerical values (deaths and hospitalizations in this case) and compare them. If the claim asserts a difference and the numbers support that difference, classify as 'supports'. If the claim asserts the opposite of the numerical difference, classify as 'refutes'. Pay close attention to the wording of the claim (e.g., 'much lower', 'higher', 'equal') to ensure accurate comparison.\textbackslash{}n\textbackslash{}nWhen evaluating claims about attendance changes, directly compare the attendance numbers for the specified dates. If the attendance on the second date is significantly lower than the first, classify the claim as 'supports'. Be mindful of the wording of the claim (e.g., 'significant decline') and ensure the numerical difference aligns with that wording. Prioritize numerical comparison over any other contextual factors.\textbackslash{}n\textbackslash{}nWhen evaluating claims about recovered footage, carefully examine the 'missing episodes with recovered footage' column for the relevant story and episode. If the table indicates that episode 4 for story 032 in Australia has recovered footage, classify the claim as 'supports'. Prioritize direct matches between the claim and the table data, especially when the claim explicitly mentions a specific episode and story number.
\\
\bottomrule
\end{tabular}
}
\caption{Baseline and optimized instructions for the Gemma3-12B model with direct prompting using various optimizers.}
\label{tab:inst_gemma12b_direct}
\end{table*}

\begin{table*}[]
\small
\resizebox{\linewidth}{!}{
\begin{tabular}{lp{0.865\textwidth}}
\toprule
\textbf{Optimizer} & \textbf{Instruction} \\
\midrule
Baseline & Verify the given claim against the provided table data. \\
\midrule
COPRO & You are an expert fact-checker specializing in data-driven verification. Your task is to rigorously evaluate a given claim against the information presented in a provided table. Determine whether the claim is definitively supported, definitively contradicted, or if the table contains no relevant information about the claim (not mentioned).  Provide a concise but thorough justification for your assessment, always referencing specific data points from the table to substantiate your reasoning. Prioritize clarity and precision in your explanation. \\
\midrule
MiPROv2 & You are an expert claim verifier. Analyze the provided claim and the accompanying table to determine if the claim is supported, refuted, or if there is not enough information to verify it. Provide a detailed, step-by-step reasoning process explaining how you arrived at your conclusion. Clearly state your final answer as \textbackslash{}"supports,\textbackslash{}" \textbackslash{}"refutes,\textbackslash{}" or \textbackslash{}"not enough info. \\
\midrule
SIMBA & Verify the given claim against the provided table data.\textbackslash{}n\textbackslash{}nWhen verifying a claim against a table, carefully examine the table's contents and units. Do not attempt calculations or comparisons that are not explicitly provided in the table. If the table lacks the necessary information to directly support or refute the claim, respond with 'not enough info'. Avoid making assumptions or performing calculations that are not directly supported by the data.\textbackslash{}n\textbackslash{}nWhen analyzing claims about performance improvements based on table data, especially when the table includes ROUGE scores, carefully consider any accompanying captions or descriptions. If the caption indicates that the system being evaluated optimizes for a different metric than the one being compared (e.g., optimizing for learned reward instead of ROUGE), do not assume that higher ROUGE scores automatically imply a performance boost. Instead, compare the system's scores to those of other systems, taking into account the optimization strategy. If the table lacks clear evidence of a performance boost relative to other systems, respond with 'not enough info'.\textbackslash{}n\textbackslash{}nWhen analyzing claims about accuracy based on table data, first identify all models relevant to the claim (e.g., models using TVMAX). Then, systematically compare the 'Test-Standard Overall' scores (or the relevant accuracy metric) for each of these models to determine which achieves the highest score. Avoid prematurely concluding that no model meets the claim's criteria before completing this comparison.\textbackslash{}n\textbackslash{}nWhen analyzing claims about energy prices, focus on the numerical values provided in the table. The claim likely refers to the price itself, not a count of prices. Carefully compare the relevant prices mentioned in the claim with the values in the table, ignoring irrelevant details like the time or percentage change. If the claim asks for a comparison between two specific items (e.g., WTI crude and natural gas), only consider those items when making your determination.
\\
\bottomrule
\end{tabular}
}
\caption{Baseline and optimized instructions for the Gemma3-12B model with CoT method using various optimizers.}
\label{tab:inst_gemma12b_cot}
\end{table*}

\begin{table*}[]
\small
\resizebox{\linewidth}{!}{
\begin{tabular}{lp{0.865\textwidth}}
\toprule
\textbf{Optimizer} & \textbf{Instruction} \\
\midrule
Baseline & Verify the given claim against the provided table data. \\
\midrule
COPRO & You are a claim verification agent. You are given a claim, a table, and a caption. Your task is to determine if the claim is supported, contradicted, or not mentioned in the table. Analyze the table data, considering the caption for context. Provide a concise answer indicating whether the claim is supported, contradicted, or not mentioned. If the claim is supported or contradicted, briefly explain how the table data supports or contradicts the claim. \\
\midrule
MiPROv2 & Verify the given claim against the provided table data.\textbackslash{}n\textbackslash{}nYou are an Agent. In each episode, you will be given the fields `claim`, `table`, `caption` as input. And you can see your past trajectory so far.\textbackslash{}nYour goal is to use one or more of the supplied tools to collect any necessary information for producing `answer`.\textbackslash{}n\textbackslash{}nTo do this, you will interleave next\_thought, next\_tool\_name, and next\_tool\_args in each turn, and also when finishing the task.\textbackslash{}nAfter each tool call, you receive a resulting observation, which gets appended to your trajectory.\textbackslash{}n\textbackslash{}nWhen writing next\_thought, you may reason about the current situation and plan for future steps.\textbackslash{}nWhen selecting the next\_tool\_name and its next\_tool\_args, the tool must be one of:\textbackslash{}n\textbackslash{}n(1) execute\_sql, whose description is \textless{}desc\textgreater      Execute a SQL query on a table provided as list of lists format.            Args:          table\_data: List of lists where first row contains headers, subsequent rows contain data          table\_name: Name of the table dataframe for executing SQL query          sql\_query: SQL query string to execute            Returns:          Formatted string of the transformed table or error message      \textless{}/desc\textgreater{}. It takes arguments \{'table\_data': \{'items': \{\}, 'type': 'array'\}, 'table\_name': \{'type': 'string'\}, 'sql\_query': \{'type': 'string'\}\}.\textbackslash{}n(2) finish, whose description is \textless{}desc\textgreater{}Marks the task as complete. That is, signals that all information for producing the outputs, i.e. `answer`, are now available to be extracted.\textless{}/desc\textgreater{}. It takes arguments \{\}.\textbackslash{}nWhen providing `next\_tool\_args`, the value inside the field must be in JSON format \\
\midrule
SIMBA & Verify the given claim against the provided table data.\textbackslash{}n\textbackslash{}nYou are an Agent. In each episode, you will be given the fields `claim`, `table`, `caption` as input. And you can see your past trajectory so far.\textbackslash{}nYour goal is to use one or more of the supplied tools to collect any necessary information for producing `answer`.\textbackslash{}n\textbackslash{}nTo do this, you will interleave next\_thought, next\_tool\_name, and next\_tool\_args in each turn, and also when finishing the task.\textbackslash{}nAfter each tool call, you receive a resulting observation, which gets appended to your trajectory.\textbackslash{}n\textbackslash{}nWhen writing next\_thought, you may reason about the current situation and plan for future steps.\textbackslash{}nWhen selecting the next\_tool\_name and its next\_tool\_args, the tool must be one of:\textbackslash{}n\textbackslash{}n(1) execute\_sql, whose description is \textless{}desc\textgreater      Execute a SQL query on a table provided as list of lists format.            Args:          table\_data: List of lists where first row contains headers, subsequent rows contain data          table\_name: Name of the table dataframe for executing SQL query          sql\_query: SQL query string to execute            Returns:          Formatted string of the transformed table or error message      \textless{}/desc\textgreater{}. It takes arguments \{'table\_data': \{'items': \{\}, 'type': 'array'\}, 'table\_name': \{'type': 'string'\}, 'sql\_query': \{'type': 'string'\}\}.\textbackslash{}n(2) finish, whose description is \textless{}desc\textgreater{}Marks the task as complete. That is, signals that all information for producing the outputs, i.e. `answer`, are now available to be extracted.\textless{}/desc\textgreater{}. It takes arguments \{\}.\textbackslash{}nWhen providing `next\_tool\_args`, the value inside the field must be in JSON format\textbackslash{}n\textbackslash{}nWhen attempting to use the `execute\_sql` tool, carefully review the tool's documentation to ensure the `table\_data` is formatted as a list of lists, where the first inner list represents the header row and subsequent lists represent the data rows. Before calling the tool, double-check the structure of `table\_data` to confirm it adheres to this format. If repeated attempts to execute the SQL query fail, abandon the tool and rely on direct analysis of the table data provided in the initial input.
\\
\bottomrule
\end{tabular}
}
\caption{Baseline and optimized instructions for the Gemma3-12B model with ReAct method using various optimizers.}
\label{tab:inst_gemma12b_react}
\end{table*}

\begin{table*}[]
\small
\resizebox{\linewidth}{!}{
\begin{tabular}{lp{0.865\textwidth}}
\toprule
\textbf{Optimizer} & \textbf{Instruction} \\
\midrule
Baseline & Verify the given claim against the provided table data. \\
\midrule
COPRO & You are a claim verifier, expertly analyzing claims against tabular data. Given a claim, a table, and a caption, you must determine if the claim is 'True', 'False', or 'Insufficient information'. Focus on precision and accuracy. Consider all relevant columns and rows. Your response *must* include the reasoning process and the final determination. Start by briefly outlining your strategy, then analyze the data, and finally state the conclusion. Only provide the determination after thorough analysis. \\
\midrule
MiPROv2 & You are an expert data analyst tasked with verifying claims against tabular data. You will be given a claim, a table, and its caption. Your objective is to analyze the table and determine if the claim is supported, refuted, or if there is not enough information to verify it.\textbackslash{}n\textbackslash{}nTo accomplish this, you will first generate Python code to extract and analyze the relevant data from the table. The generated code should be printed to the console within a fenced code block. This code may involve parsing the table, calculating statistics, or performing comparisons.\textbackslash{}n\textbackslash{}nAfter generating the code, execute it and observe the output. Based on the output and your understanding of the claim and table, determine whether the claim is supported, refuted, or if there is not enough information.\textbackslash{}n\textbackslash{}nFinally, mark `finished=True` along with the final Python code. Ensure all necessary information is extracted before marking `finished=True`.\textbackslash{}n\textbackslash{}nYou have access to the Python Standard Library. \\
\midrule
SIMBA & Verify the given claim against the provided table data.\textbackslash{}n\textbackslash{}nYou are an intelligent agent. For each episode, you will receive the fields `claim`, `table`, `caption` as input.\textbackslash{}nYour goal is to generate executable Python code that collects any necessary information for producing `answer`.\textbackslash{}nFor each iteration, you will generate a code snippet that either solves the task or progresses towards the solution.\textbackslash{}nEnsure any output you wish to extract from the code is printed to the console. The code should be enclosed in a fenced code block.\textbackslash{}nWhen all information for producing the outputs (`answer`) are available to be extracted, mark `finished=True` besides the final Python code.\textbackslash{}nYou have access to the Python Standard Library and the following functions:\textbackslash{}n\textbackslash{}nWhen receiving the `table` input, which is a string containing tabular data, use the `pandas` library to parse the table data directly from the string. Specifically, use `pd.read\_csv` with `StringIO` to read the table string into a pandas DataFrame. This will allow you to access the table data using DataFrame indexing and avoid the `NameError` that occurs when trying to access an undefined variable.
\\
\bottomrule
\end{tabular}
}
\caption{Baseline and optimized instructions for the Gemma3-12B model with CodeAct method using various optimizers.}
\label{tab:inst_gemma12b_codeact}
\end{table*}

\begin{table*}[]
\small
\resizebox{\linewidth}{!}{
\begin{tabular}{lp{0.865\textwidth}}
\toprule
\textbf{Optimizer} & \textbf{Instruction} \\
\midrule
Baseline & Verify the given claim against the provided table data. \\
\midrule
COPRO & You are a fact-checking assistant. Your task is to determine if a given claim is supported, refuted, or not found within a provided data table.\textbackslash{}n\textbackslash{}nHere's how you'll proceed:\textbackslash{}n\textbackslash{}n1. **Carefully read the claim.** Understand exactly what it states.\textbackslash{}n2. **Analyze the data table.** Identify relevant rows and columns to assess the claim.\textbackslash{}n3. **Determine the relationship:**\textbackslash{}n   * **Supported:** If the data in the table directly confirms the claim.\textbackslash{}n   * **Refuted:** If the data in the table directly contradicts the claim.\textbackslash{}n   * **Not found:** If the table does not contain information relevant to the claim, or the information is insufficient to verify it.\textbackslash{}n4. **Respond concisely with *only* one of the following labels:** `Supported`, `Refuted`, or `Not found`. Do not include explanations or additional text. \\
\midrule
MiPROv2 & You are an expert at fact verification. Given a claim, a table, and a caption describing the table, determine if the claim is supported by the table, refuted by the table, or if the table provides insufficient information to make a determination.\textbackslash{}n\textbackslash{}nCarefully analyze the claim and the table content. Pay attention to units, specific data points, and the overall context. Your response should be one of the following: 'supports', 'refutes', or 'not enough info'.\textbackslash{}n\textbackslash{}nHere's an example:\textbackslash{}n\textbackslash{}nClaim: story number 032 from australia has episode 4 as the missing episode with recovered footage\textbackslash{}nTable: || doctor | season | story no | serial | number of episodes | total footage remaining from missing episodes (mm : ss) | missing episodes with recovered footage | country / territory | source | format | total footage (mm : ss) || ...\textbackslash{}nCaption: doctor who missing episodes\textbackslash{}nAnswer: supports\textbackslash{}n\textbackslash{}nNow, apply your expertise to the given claim, table, and caption. \\
\midrule
SIMBA & Verify the given claim against the provided table data.\textbackslash{}n\textbackslash{}nWhen presented with a claim and a table, focus on directly comparing the values in the table to the statement in the claim. In this case, the claim states that 'adding logits' does *not* perform the best. The table shows 'Add Logits' achieving 80.85 F1\%, 'Add Logits+Expert' achieving 80.90 F1\%, and other methods having lower or similar F1 scores. Therefore, the claim is refuted because adding logits *does* perform well, and is not the worst. If you encounter similar claims, explicitly check if the table data contradicts the claim's assertion.\textbackslash{}n\textbackslash{}nWhen evaluating a claim about the effectiveness of an approach based on table data, prioritize identifying the performance metrics associated with that approach. If the metrics demonstrate strong performance (e.g., high success rates, low error rates), the claim that the approach is *not* effective is likely refuted. Conversely, if the metrics show poor performance, the claim is supported. Pay close attention to comparative values within the table to determine if the approach outperforms others, even if absolute values are not exceptionally high.\textbackslash{}n\textbackslash{}nWhen you receive a claim and a table with numerical data, explicitly compare the numbers in the claim to the numbers in the table. If the numbers in the table demonstrably contradict the claim (e.g., the claim states values are within a range, but the table shows values outside that range), then output 'refutes'. Do not default to 'not enough info' if numerical comparisons can be made.\textbackslash{}n\textbackslash{}nWhen you receive a claim and a table with numerical data, *always* carefully read the caption to understand what the numbers represent and how they relate to the claim. If the caption indicates that the numbers are not directly comparable to the claim's assertion (e.g., different metrics are being optimized), or if the claim refers to a factor not represented in the table, output 'not enough info'. Do not attempt to make a judgment based solely on numerical comparisons if the context suggests it's inappropriate.
\\
\bottomrule
\end{tabular}
}
\caption{Baseline and optimized instructions for the Gemma3-27B model with direct prompting using various optimizers.}
\label{tab:inst_gemma27b_direct}
\end{table*}

\begin{table*}[]
\small
\resizebox{\linewidth}{!}{
\begin{tabular}{lp{0.865\textwidth}}
\toprule
\textbf{Optimizer} & \textbf{Instruction} \\
\midrule
Baseline & Verify the given claim against the provided table data. \\
\midrule
COPRO & You are a meticulous fact-checker. Your task is to determine if a given claim is supported by the data in a provided table.\textbackslash{}n\textbackslash{}nFirst, carefully analyze the claim and identify the key pieces of information needed to verify it. Then, examine the table to find relevant data.\textbackslash{}n\textbackslash{}nBased on this comparison, respond with one of the following options:\textbackslash{}n\textbackslash{}n* **Supported:** If the table data directly supports the claim.\textbackslash{}n* **Contradicted:** If the table data directly contradicts the claim.\textbackslash{}n* **Not enough information:** If the table does not contain enough information to verify the claim.\textbackslash{}n\textbackslash{}nPresent your reasoning *before* your final answer. Explain how you arrived at your conclusion by referencing specific data points from the table. \\
\midrule
MiPROv2 & You are a highly skilled fact-checker responsible for ensuring the accuracy of critical information used in public health advisories. A misclassification could have serious consequences. Given a claim, a table of data, and a caption describing the table, your task is to meticulously verify the claim against the data. Provide a step-by-step reasoning process, clearly outlining how you arrived at your conclusion. Then, definitively state whether the claim is 'supports', 'refutes', or 'not enough info' based on the table. Your accuracy is paramount. \\
\midrule
SIMBA & Verify the given claim against the provided table data.\textbackslash{}n\textbackslash{}nWhen analyzing claims about statistical significance, prioritize identifying whether the table explicitly indicates statistical significance (e.g., using asterisks or other markers) for the relevant models and conditions. If the table shows statistically significant improvements for a model listed as 'Our Models' under oracle setup, even without a direct comparison to the same model without oracle setup, the claim that 'our model does not achieve statistically significantly better BLEU scores' is likely refuted.\textbackslash{}n\textbackslash{}nWhen analyzing claims, focus *strictly* on whether the table contains information directly relevant to the claim. Avoid making inferences or drawing conclusions based on related data (like accuracy scores in this case) if the claim concerns a different aspect (like the use of sense-level information). If the table doesn't explicitly address the claim, default to 'not enough info', even if you can reason about related concepts.\textbackslash{}n\textbackslash{}nWhen analyzing claims, strictly adhere to the information presented in the table. If the table defines ranges or boundaries (like 'less than 17\%'), avoid interpreting the claim as being 'inaccurate' simply because a value falls within that range. Instead, focus on whether the table *directly* supports, refutes, or lacks information regarding the claim. If the table only provides definitions or categorizations without addressing the claim's truthfulness, default to 'not enough info'.
\\
\bottomrule
\end{tabular}
}
\caption{Baseline and optimized instructions for the Gemma3-27B model with CoT method using various optimizers.}
\label{tab:inst_gemma27b_cot}
\end{table*}

\begin{table*}[]
\small
\resizebox{\linewidth}{!}{
\begin{tabular}{lp{0.865\textwidth}}
\toprule
\textbf{Optimizer} & \textbf{Instruction} \\
\midrule
Baseline & Verify the given claim against the provided table data. \\
\midrule
COPRO & You are a highly accurate claim verification agent. Your task is to determine whether a given claim is supported by data within a provided table. You will be given a `claim`, a `table` (in list of lists format including a header row), and a `caption` describing the table.\textbackslash{}n\textbackslash{}nFollow these steps:\textbackslash{}n\textbackslash{}n1. **Understand the Claim:** Carefully read the `claim` to identify the key entities and relationship being asserted.\textbackslash{}n2. **Analyze the Table:** Examine the `table` structure, header, and data to understand the information it contains. Consider the `caption` for helpful context.\textbackslash{}n3. **Formulate a SQL Query:** Construct a SQL query that, when executed against the `table`, will directly answer the question posed by the `claim`. The SQL query should be as simple and direct as possible.\textbackslash{}n4. **Execute the SQL Query:** Use the `execute\_sql` tool with the `table\_data`, a chosen `table\_name` (e.g., \textbackslash{}"my\_table\textbackslash{}"), and your crafted `sql\_query`.\textbackslash{}n5. **Interpret the Results:** Analyze the results returned by the `execute\_sql` tool. \textbackslash{}n6. **Verify the Claim:** Based on the query results, determine if the claim is supported (TRUE), contradicted (FALSE), or if the table does not contain sufficient information to determine (UNKNOWN).\textbackslash{}n7. **Finish:** Use the `finish` tool to signal task completion.\textbackslash{}n\textbackslash{}nYour response should be accurate and concise, relying solely on the information presented in the `table`. Prioritize SQL queries that directly address the claim's core assertion. \\
\midrule
MiPROv2 & You are a fact-checking agent tasked with verifying claims against tabular data. You will be given a claim, a table, and a table caption. Your goal is to determine if the table supports, refutes, or provides insufficient information to verify the claim.\textbackslash{}n\textbackslash{}n**Here's how you should operate:**\textbackslash{}n\textbackslash{}n1. **Reasoning:** Carefully read the claim, table caption, and table data. Formulate a clear plan to determine the answer. Consider potential issues like units, data types, and ambiguous language.\textbackslash{}n2. **Tool Use (execute\_sql):** If the table is complex or requires specific data extraction, use the `execute\_sql` tool to query the table.  Craft SQL queries that directly address the claim.  Remember to handle potential errors (e.g., invalid column names, data types).  If SQL fails, proceed to manual inspection.\textbackslash{}n3. **Manual Inspection:** If `execute\_sql` fails or isn't necessary, manually inspect the table to find relevant information.\textbackslash{}n4. **Final Decision:** Based on your reasoning and the extracted evidence, determine whether the table *supports* the claim, *refutes* the claim, or provides *not enough info* to make a determination.\textbackslash{}n5. **Output:**  Provide your reasoning step-by-step, then state your final answer as either 'supports', 'refutes', or 'not enough info'.\textbackslash{}n\textbackslash{}n**Important Considerations:**\textbackslash{}n\textbackslash{}n*   Prioritize accuracy. Double-check your reasoning and evidence.\textbackslash{}n*   Be mindful of the table caption; it often provides crucial context.\textbackslash{}n*   If the claim refers to a comparison, ensure you are comparing the correct data points.\textbackslash{}n*   If the table does not contain the information necessary to answer the claim, select \textbackslash{}"not enough info\textbackslash{}". \\
\midrule
SIMBA & Verify the given claim against the provided table data.\textbackslash{}n\textbackslash{}nYou are an Agent. In each episode, you will be given the fields `claim`, `table`, `caption` as input. And you can see your past trajectory so far.\textbackslash{}nYour goal is to use one or more of the supplied tools to collect any necessary information for producing `answer`.\textbackslash{}n\textbackslash{}nTo do this, you will interleave next\_thought, next\_tool\_name, and next\_tool\_args in each turn, and also when finishing the task.\textbackslash{}nAfter each tool call, you receive a resulting observation, which gets appended to your trajectory.\textbackslash{}n\textbackslash{}nWhen writing next\_thought, you may reason about the current situation and plan for future steps.\textbackslash{}nWhen selecting the next\_tool\_name and its next\_tool\_args, the tool must be one of:\textbackslash{}n\textbackslash{}n(1) execute\_sql, whose description is \textless{}desc\textgreater      Execute a SQL query on a table provided as list of lists format.            Args:          table\_data: List of lists where first row contains headers, subsequent rows contain data          table\_name: Name of the table dataframe for executing SQL query          sql\_query: SQL query string to execute            Returns:          Formatted string of the transformed table or error message      \textless{}/desc\textgreater{}. It takes arguments \{'table\_data': \{'items': \{\}, 'type': 'array'\}, 'table\_name': \{'type': 'string'\}, 'sql\_query': \{'type': 'string'\}\}.\textbackslash{}n(2) finish, whose description is \textless{}desc\textgreater{}Marks the task as complete. That is, signals that all information for producing the outputs, i.e. `answer`, are now available to be extracted.\textless{}/desc\textgreater{}. It takes arguments \{\}.\textbackslash{}nWhen providing `next\_tool\_args`, the value inside the field must be in JSON format\textbackslash{}n\textbackslash{}nWhen you receive table data directly as a list of lists, avoid attempting to use the `execute\_sql` tool. Instead, directly analyze the provided data to answer the question. Prioritize direct analysis of the provided data before resorting to SQL queries, especially if initial SQL attempts fail.\textbackslash{}n\textbackslash{}nWhen the table data is provided directly as a list of lists, prioritize analyzing the data directly instead of attempting to use SQL. SQL is useful for complex queries on large datasets, but for small, directly provided tables, direct analysis is more efficient and avoids potential errors. Look for keywords like 'list of lists' or 'table' in the input to determine if direct analysis is appropriate.\textbackslash{}n\textbackslash{}nWhen encountering an Odds Ratio (OR) of 1.00, especially in the context of a claim about a baseline percentage, consider explicitly stating the assumption that the baseline rate is equal to the claimed percentage. Avoid immediately concluding 'not enough info' simply because the table doesn't directly state the percentage; instead, explore reasonable interpretations of the OR value in relation to the claim.\textbackslash{}n\textbackslash{}nWhen analyzing tables, prioritize the most directly relevant value to the claim. In this case, the claim asks for the sensitivity *percentage* of the acute phase. While 98\% appears in the table, it's a specificity value, not a sensitivity value. Focus on the 'Sensitivity (\%)' column and the 'Acute phase' row to find the correct value (17\%). Avoid getting distracted by other numbers in the table that aren't directly responsive to the question.\textbackslash{}n\textbackslash{}nWhen the input `table` is provided as a list of lists (rather than a database connection or similar), prioritize direct analysis of the data instead of attempting to use SQL. SQL is more appropriate for large datasets or complex queries, but for small, directly provided tables, direct analysis is more efficient and avoids potential errors. Look for keywords like 'list of lists' or 'table' in the input to determine if direct analysis is appropriate. If SQL queries repeatedly fail, immediately switch to direct data analysis.
\\
\bottomrule
\end{tabular}
}
\caption{Baseline and optimized instructions for the Gemma3-27B model with ReAct method using various optimizers.}
\label{tab:inst_gemma27b_react}
\end{table*}

\begin{table*}[]
\small
\resizebox{\linewidth}{!}{
\begin{tabular}{lp{0.865\textwidth}}
\toprule
\textbf{Optimizer} & \textbf{Instruction} \\
\midrule
Baseline & Verify the given claim against the provided table data. \\
\midrule
COPRO & You are a highly skilled agent designed to verify claims based on data presented in tables. You will be given a `claim`, a `table` (represented as a string), and a `caption` describing the table. Your task is to determine whether the claim is supported by the information in the table. \textbackslash{}n\textbackslash{}nHere's how you will proceed:\textbackslash{}n\textbackslash{}n1. **Understand the Table:** Parse the table string to understand its structure (rows, columns, headers).\textbackslash{}n2. **Extract Relevant Data:** Identify the data within the table that relates to the `claim`.\textbackslash{}n3. **Verify the Claim:** Compare the information in the `claim` with the extracted data to determine if the claim is TRUE, FALSE, or UNKNOWN (if the table doesn't contain enough information to verify the claim).\textbackslash{}n4. **Generate Python Code:**  Write Python code to accomplish steps 1-3. The code must print the final answer (`TRUE`, `FALSE`, or `UNKNOWN`) to standard output.  Focus on extracting and comparing the key pieces of information needed to answer the question directly.\textbackslash{}n5. **Mark Completion:**  Include `finished=True` after the final code block when the answer is ready to be extracted.\textbackslash{}n\textbackslash{}nDo not include any conversational text or explanations in your output; only the Python code and the `finished=True` marker. \\
\midrule
MiPROv2 & You are a highly skilled data analyst tasked with verifying claims against tabular data. You will receive a `claim` (a statement to be verified), a `table` (containing the data in a delimited string format), and a `caption` (describing the table). Your objective is to determine if the `claim` is supported, refuted, or if the `table` provides insufficient information (`not enough info`).\textbackslash{}n\textbackslash{}nYou must generate Python code to parse the `table` data using the `pandas` library. The table data is delimited by `||`. Extract the relevant information from the table to evaluate the claim.  Ensure your code handles potential errors gracefully, such as missing data or incorrect data types. Print the final boolean result (True or False) representing whether the claim is supported by the table data to the console.\textbackslash{}n\textbackslash{}nYour response must follow these steps:\textbackslash{}n\textbackslash{}n1.  **Parse the Table:** Use `pandas` to read the table data from the provided string. Use `sep='||'` to correctly parse the data. Clean the data by removing leading/trailing whitespace from column names and cell values.\textbackslash{}n2.  **Analyze the Claim:**  Carefully read and understand the claim. Identify the specific data points needed from the table to verify the claim.\textbackslash{}n3.  **Extract and Compare:** Extract the relevant data points from the `pandas` DataFrame. Perform the necessary comparison or calculation to determine if the claim is true or false based on the table data.\textbackslash{}n4.  **Print the Result:** Print either `True` or `False` to the console, representing the result of your analysis.\textbackslash{}n5.  **Mark Completion:** After printing the result, add `Finished: True` to signify the task is complete.\textbackslash{}n\textbackslash{}nYou should only print the boolean value (True or False) and `Finished: True` to the console; no other text or explanations. \\
\midrule
SIMBA & Verify the given claim against the provided table data.\textbackslash{}n\textbackslash{}nYou are an intelligent agent. For each episode, you will receive the fields `claim`, `table`, `caption` as input.\textbackslash{}nYour goal is to generate executable Python code that collects any necessary information for producing `answer`.\textbackslash{}nFor each iteration, you will generate a code snippet that either solves the task or progresses towards the solution.\textbackslash{}nEnsure any output you wish to extract from the code is printed to the console. The code should be enclosed in a fenced code block.\textbackslash{}nWhen all information for producing the outputs (`answer`) are available to be extracted, mark `finished=True` besides the final Python code.\textbackslash{}nYou have access to the Python Standard Library and the following functions:\textbackslash{}n\textbackslash{}nWhen generating code to parse tables, pay close attention to syntax, especially indentation. Ensure that all `else` statements have corresponding indented blocks. Before finalizing the code, consider adding a simple test case to verify that the parsing logic works correctly with a small subset of the table data. Also, consider adding error handling to catch potential parsing issues and provide informative error messages.\textbackslash{}n\textbackslash{}nWhen parsing tables, be extremely careful with the separator. The table uses '||' as a separator, but you initially used '\textbackslash{}\textbackslash{}|'. Ensure the separator matches the table's format exactly. Also, double-check the column names you are trying to access after parsing the table to avoid KeyErrors. Consider adding a small test case to verify the parsing logic with a subset of the table data before proceeding with the full analysis.\textbackslash{}n\textbackslash{}nWhen parsing the table, double-check the separator used in `pd.read\_csv`. The table uses '||' as a separator. Ensure this is correctly specified. Also, after creating the DataFrame, verify the number of columns and the assigned column names to ensure they match the table structure. Add a small test case to verify the parsing logic with a subset of the table data before proceeding with the full analysis. If parsing fails, return a simple error message instead of crashing.\textbackslash{}n\textbackslash{}nWhen generating Python code, pay extremely close attention to indentation. Incorrect indentation is a common source of errors. Before finalizing the code, use a linter or run the code locally to catch these errors. Specifically, ensure that all blocks following control flow statements (e.g., `if`, `else`, `try`, `except`) are properly indented.
\\
\bottomrule
\end{tabular}  
}
\caption{Baseline and optimized instructions for the Gemma3-27B model with CodeAct method using various optimizers.}
\label{tab:inst_gemma27b_codeact}
\end{table*}  

\end{document}